\documentclass[review,authoryear]{elsarticle}
\graphicspath{ {./figures/} }

\usepackage{hyperref}
\usepackage{float}
\usepackage{verbatim}   % provides the comment environment
\IfFileExists{apalike.sty}{\usepackage{apalike}}{}

\usepackage{amsmath,amssymb,amsfonts}
\usepackage{graphicx}
\usepackage{booktabs}
\usepackage{multirow}
\usepackage{xcolor}
\usepackage{geometry}
\usepackage{algorithm,algpseudocode}
\usepackage{tabularx}
\usepackage{bbm}
\usepackage{threeparttable}
\usepackage{makecell}   % multi-line cells
\newcommand{\etal}{et~al. }

\restylefloat{figure}
\floatstyle{plaintop}   % table caption at top
\restylefloat{table}

\journal{Expert Systems with Applications}

\begin{document}
\begin{frontmatter}

%% ---- Standalone title page required by the ESWA template ----
\begin{titlepage}
\begin{center}
\vspace*{1cm}

\textbf{\Large Hierarchical Building–Floor–Room Framework for Privacy-Preserving Autonomous Delivery in Large-Scale Buildings}

\vspace{1.5cm}

% Only the two corresponding authors' e-mail addresses and the three
% affiliation addresses were available in A/main.tex. Missing items are
% flagged rather than fabricated. ESWA additionally requires full postal
% addresses (street, postcode); please complete the three lines below.
XU Xiaoji$^{a}$ (xxuco@connect.ust.hk), CHEN Jinhua$^{c}$ (jchenjb@connect.ust.hk),
Zhang Leihan$^{b}$ (lzhang305@connect.hkust-gz.edu.cn), LUO Rui$^{b}$ (rluo041@connect.hkust-gz.edu.cn),
LIAO Zhongyuan$^{c}$ (zy.liao@connect.ust.hk) \\

\vspace{10pt}

\begin{flushleft}
\small
$^a$ Division of Integrative Systems and Design, Hong Kong University of Science and Technology, Hong Kong, China \\
$^b$ Hong Kong University of Science and Technology (Guangzhou), Guangzhou, China \\
$^c$ Hong Kong University of Science and Technology, Hong Kong, China

\vspace{1cm}
\textbf{Corresponding authors at:} \\

LUO Rui \\
Hong Kong University of Science and Technology (Guangzhou), Guangzhou, China \\
Email: rluo041@connect.hkust-gz.edu.cn \\[6pt]

LIAO Zhongyuan \\
Hong Kong University of Science and Technology, Hong Kong, China \\
Email: zy.liao@connect.ust.hk

\end{flushleft}
\end{center}
\end{titlepage}

\title{Hierarchical Building–Floor–Room Framework for Privacy-Preserving Autonomous Delivery in Large-Scale Buildings}

%% ---- Authors, e-mails, corresponding-author note and affiliations ----
\author[label2]{XU Xiaoji}
\author[label2]{CHEN Jinhua}

\author[label3]{Zhang Leihan}

\author[label3]{LUO Rui\corref{cor1}}
\ead{rluo041@connect.hkust-gz.edu.cn}
\author[label1]{LIAO Zhongyuan\corref{cor1}}
\ead{zy.liao@connect.ust.hk}
%% Corresponding authors
\cortext[cor1]{Corresponding authors.}

%% Addresses
\address[label2]{Division of Integrative Systems and Design, Hong Kong University of Science and Technology, Hong Kong, China}
\address[label3]{Hong Kong University of Science and Technology (Guangzhou), Guangzhou, China}
\address[label1]{Hong Kong University of Science and Technology, Hong Kong, China}

%% ---- Abstract ----
\begin{abstract}
Autonomous delivery in large-scale buildings requires coordinated planning across multiple spatial levels, including building, floor, and room, while ensuring scheduling efficiency. Existing methods often focus solely on algorithmic development, neglecting integration with physical robotic platforms, and rarely validate performance in real-world scenarios. Many approaches also overlook hierarchical building structures and room-level privacy constraints, limiting practical applicability. To address these challenges, this paper presents a hierarchical delivery framework based on a Building–Floor–Room abstraction, which decomposes the problem across multiple levels to significantly reduce optimization complexity. We formulate a building-aware extension of the capacitated vehicle routing problem that explicitly models inter-building and inter-floor connectivity, room-level access constraints, and privacy-sensitive areas, enabling globally feasible and privacy-compliant task assignment and routing. At the execution level, we introduce a novel hardware design for non-interactive, privacy-preserving delivery, supported by reachability-aware distances and a closed-loop mapping, localization, and navigation system, ensuring consistency between planned routes and physical navigation. The proposed framework is validated on a real multi-sensor robotic platform, demonstrating reliable multi-floor delivery with improved path rationality and overall efficiency.
\end{abstract}

%% ---- Keywords ----

\end{frontmatter}

\section{Introduction}

The increasing demand for last-mile delivery in building complexes, campuses, and large residential communities necessitates mobile robots capable of executing multi-goal delivery tasks in real-world indoor environments~\citep{tilk2021last}. Unlike conventional indoor navigation, which often focuses on a single floor or building, real-world facilities exhibit a \textbf{three-level hierarchical structure}—\textbf{building, floor, and room}—each posing distinct operational challenges~\citep{rosinol2021kimera}:
(1) \emph{Building-level}, where physical separation and sparse inter-building connections make inter-building travel costly and planning-sensitive;
(2) \emph{Floor-level}, where vertical mobility relies on elevators with vertical transfer cost and nontrivial waiting times, creating significant scheduling constraints;
(3) \emph{Room-level}, where each floor features complex 2D topologies, branched corridors, diverse room layouts, and privacy-restricted zones, complicating local path planning and reachability.

Such hierarchical structures pose fundamental challenges to classical Vehicle Routing Problem (VRP) formulations. VRP typically relies on ideal Euclidean distances and metric properties (e.g., symmetry and triangle inequality), assumptions fundamentally violated by non-Euclidean path distances, elevator access costs, and time-varying vertical transfers. Furthermore, while the base Capacitated VRP (CVRP) is already known to be NP-hard \citep{archetti2008split}, constructing a unified graph spanning all building, floor, and room levels exacerbates this complexity. This unified modeling leads to a combinatorial explosion of the state space, rendering the global hierarchical routing problem intractable for real-time operation and preventing timely responses to dynamic delivery requests~\citep{fragapane2021planning}.

Beyond structural complexity, reliable multi-goal delivery in real-world deployments is further hindered by three key practical challenges.

\textbf{(1) Lack of effective large-scale cross-building optimization models.}
Existing approaches focus on single-building problems, and do not scale to delivery scenarios involving multiple buildings and heterogeneous inter-building distances. As a result, cross-building delivery decisions are often handled by rules (e.g., first-in-first-out), leading to inefficient capacity utilization and excessive depot returns.

\textbf{(2) Dynamic environments and sensing limitations.}
Real-world indoor environments are highly dynamic, with moving pedestrians and time-varying layouts that disrupt stable perception and localization. These factors introduce uncertainty into navigation and execution, necessitating robust odometry and sensor fusion methods to maintain reliable operation~\citep{xu2021fastlio}.

\textbf{(3) Plan--execution mismatch in decoupled systems.}
Global VRP solutions are typically computed on idealized maps and assume perfect execution. In practice, however, discrepancies between planned routes and feasible robot trajectories frequently arise. Moreover, mapping, localization, scheduling, and navigation modules are often designed and optimized independently, preventing closed-loop coordination and resulting in inconsistent end-to-end performance.

% Moreover, with the increasing adoption of autonomous delivery systems and the rising cost of manual labor, delivery efficiency has become an important consideration in residential building design. In many communities, building layout and depot placement directly affect the feasibility and performance of unmanned delivery. Therefore, it is also essential to use quantitative models to guide building layout design toward better support for autonomous delivery.

To address these challenges, a \textbf{hierarchical autonomous delivery system} tailored to the Building--Floor--Room structure is proposed. The main contributions of this work are summarized as follows:

\begin{itemize}
    \item A \textbf{hierarchical Building--Floor--Room model} and a \textbf{building-aware CVRP formulation} are introduced, accounting for inter-building and inter-floor connectivity, room-level access, and privacy-sensitive areas, which reduces optimization complexity while enabling globally feasible and privacy-compliant routing.
    
    \item A \textbf{unified mapping and autonomous navigation system} integrated with a \textbf{non-interactive, privacy-preserving robot design} is developed, ensuring consistency between hierarchical planning and real-world execution.
    
    % \item A \textbf{layout evaluation and depot optimization method} is proposed, allowing quantitative assessment of delivery efficiency across different building configurations and reasonable depot placement.

    \item A \textbf{comprehensive computational and empirical evaluation} quantifying the performance of the hierarchical framework against existing baselines, demonstrating significant efficiency improvements over cross-building delivery baselines, and validating system reliability via real-world multi-building, multi-floor field deployments.

\end{itemize}

% Overall, this work provides a principled and deployable framework for autonomous delivery robots operating in large-scale, hierarchical building--floor--room environments.

%%%%%%%%%%%%%%%%%%%%%%%%%%%%%%%%%%%%%%%%%%%%%%%%%%%%%%%%%%%%%%%%%%%%%%%%%%%%%%%%

\section{Related Work}

Establishing reliable autonomous delivery in complex building complexes requires novel solutions across environmental modeling, operational research, and robotic perception. This section reviews the state-of-the-art in these three interconnected domains, highlighting the existing limitations that motivate our proposed hierarchical autonomous delivery system.

\subsection{Hierarchical and Semantic Environmental Modeling for Navigation}
Autonomous navigation fundamentally relies on accurate environment representation. 
Pioneering works established metric-based grid maps and early topological graphs~\citep{thrun1996integrating} to represent planar environments. 
Modern facilities, however, demand multi-level abstractions that can explicitly capture building-scale structure and semantics. Han et al.~\citeyearpar{Han2025_multifloor_robot_delivery} and Palacín et al.~\citeyearpar{Palacin2023_multistory_delivery_path} model robots navigating multi-floor buildings, explicitly considering elevator and stair nodes. 
To this end, hierarchical environment models and graph-based representations have been proposed to manage spatial complexity, 
ranging from extended hierarchical building graphs for design and engineering~\citep{slusarczyk2017extended}, 
to hierarchical attributed graphs for facade and building component modeling~\citep{wang2024hierarchical}, 
and indoor navigation graphs that explicitly encode multi-storey visibility and connectivity~\citep{zhou2022hivg}. 
Such hierarchical mapping structures often segment the environment into connected regions or levels~\citep{jeong2018multimodal,wang2018dels}, 
enabling scalable reasoning across floors and functional spaces. 
As environment scale increases, optimizing large unified graph structures becomes computationally expensive. 
Hierarchical or multi-level graph representations can significantly improve efficiency; however, designing effective hierarchical models that balance geometric fidelity, semantic richness, and computational tractability remains an open research problem.

With the prevalence of 3D sensing, research has moved towards dense 3D reconstruction using methods like TSDF \citep{kim2019rgb,fehr2017tsdf}, and scalable volumetric maps \citep{yang2022vox}. Concurrently, semantic mapping has gained traction, enriching geometric models with object and place labels. Recent systems, including Kimera \citep{rosinol2020kimera} and Hydra \citep{hughes2022hydra}, represent the environment using 3D Scene Graphs, unifying metric and semantic information. Gu \etal proposed ConceptGraphs, an open-vocabulary graph-structured representation for 3D scenes  \citep{gu2024conceptgraphs}. While these semantic approaches provide rich contextual representations, they mainly focus on scene understanding or geometric reconstruction. Meanwhile, with the growing emphasis on user privacy, explicitly modeling privacy-sensitive areas (e.g., restricted offices or sensitive zones) has become increasingly important. Existing privacy-preserving efforts, such as the multi-factor authentication framework proposed by Yang \etal \citeyearpar{yang2025privacy}, primarily protect identities and communication security in robotic delivery systems rather than spatial semantic representations. As a result, few semantic environment models explicitly incorporate privacy constraints into their spatial abstractions. Our work addresses this gap by integrating semantic connectivity, privacy-sensitive regions, and cost modeling into a unified Building–Floor–Room hierarchical graph optimized for delivery and navigation tasks.

\subsection{Vehicle Routing Problems with Indoor and Dynamic Constraints}
The Vehicle Routing Problem (VRP) is the core of multi-goal delivery, with foundational theory well-documented in classical works \citep{toth2014vehicle}. Standard VRP formulations rely on metric assumptions, such as symmetric and Euclidean distances, which fail dramatically in structured indoor settings \citep{siegwart2011introduction}.

Efforts to adapt VRP to robotic systems have focused on incorporating physical constraints. The VRP with Time Windows (VRP-TW) \citep{solomon1987algorithms} is relevant for timed deliveries, while stochastic VRP addresses uncertainty in service times. For indoor logistics, the primary challenge is the non-Euclidean topology and the asymmetric cost induced by vertical movement. Studies have attempted to model elevator waiting times \citep{paredes2024modeling} and integrated multi-floor scheduling with localization \citep{kim2024development}. Furthermore, optimizing for energy (Green VRP) \citep{sabet2022green} is crucial for battery-powered AMRs. From an algorithmic perspective, both metaheuristics like Ant Colony Optimization \citep{stodola2022adaptive} and emerging learning-based approaches \citep{bogyrbayeva2024machine} have been applied to VRP variants. In addition to single-robot navigation, multiple robots must be coordinated to optimize overall delivery efficiency. Multi-trip vehicle routing problems (MTVRP) are used in \cite{Han2025_multifloor_robot_delivery} to schedule multiple robots across multiple floors while minimizing total travel cost. However, existing VRP applications for robotics generally operate on idealized graphs. They lack a mechanism to dynamically incorporate SLAM-derived reachability and consistency information, leading to the persistent plan–execution mismatch when transitioning between different building levels. We overcome this by formulating a hierarchical CVRP whose distance matrix is derived directly from a reachability-aware map pipeline.

\subsection{LiDAR SLAM, Fusion, and Closed-Loop System Integration}
Reliable autonomy hinges on robust Simultaneous Localization and Mapping (SLAM). The progression from 2D methods like Cartographer \citep{hess2016real} to 3D LiDAR SLAM, starting with LOAM \citep{zhang2014loam}, has been critical. Contemporary systems like LIO-SAM \citep{shan2020lio}, Invariant-DLIO \citep{fang2025invariant}, and the highly efficient FAST-LIO2 \citep{xu2022fast} use tight LiDAR-Inertial fusion to maintain accuracy over long trajectories. For additional robustness, visual-inertial odometry (VIO), pioneered by systems like VINS-Mono \citep{qin2018vins} and DSO \citep{wang2017stereo}, and full Lidar-Visual-Inertial fusion (LVI-SAM \citep{shan2021lvi}, R3LIVE \citep{lin2022r3live}) have been developed to mitigate drift in feature-poor or geometrically degenerate environments. Graph-based SLAM systems are also critical for global consistency and loop closure \citep{koide2019portable, mur2015orb}.

Despite the sophistication of these modules, they often operate in isolation. The widely used navigation architecture consists of global planning feeding into local planners like the Dynamic Window Approach (DWA) \citep{fox1997dynamic} or the Time-Elastic Band (TEB) \citep{rosmann2017integrated}. Crucially, this setup inherently suffers from decoupling: the high-level VRP scheduler (Section 2.2) is blind to real-time SLAM consistency and execution challenges (e.g., elevator localization degradation). While theoretical work has outlined the necessity of integrated systems \citep{cadena2016past, lin2019loam}, few deployable architectures have achieved true closed-loop integration. Our work represents a significant step by implementing a closed-loop system that leverages Fast-LIO for high-fidelity state estimation and feeds real-time spatial information directly back into the hierarchical task scheduler, thereby ensuring global-local plan consistency.

%%%%%%%%%%%%%%%%%%%%%%%%%%%%%%%%%%%%%%%%%%%%%%%%%%%%%%%%%%%%%%%%%%%%%%%%%%%%%%%%

\section{System Overview and Hierarchical CVRP Modeling}
\label{sec:system_model}

\begin{figure}
    \centering
    \includegraphics[width=1\linewidth]{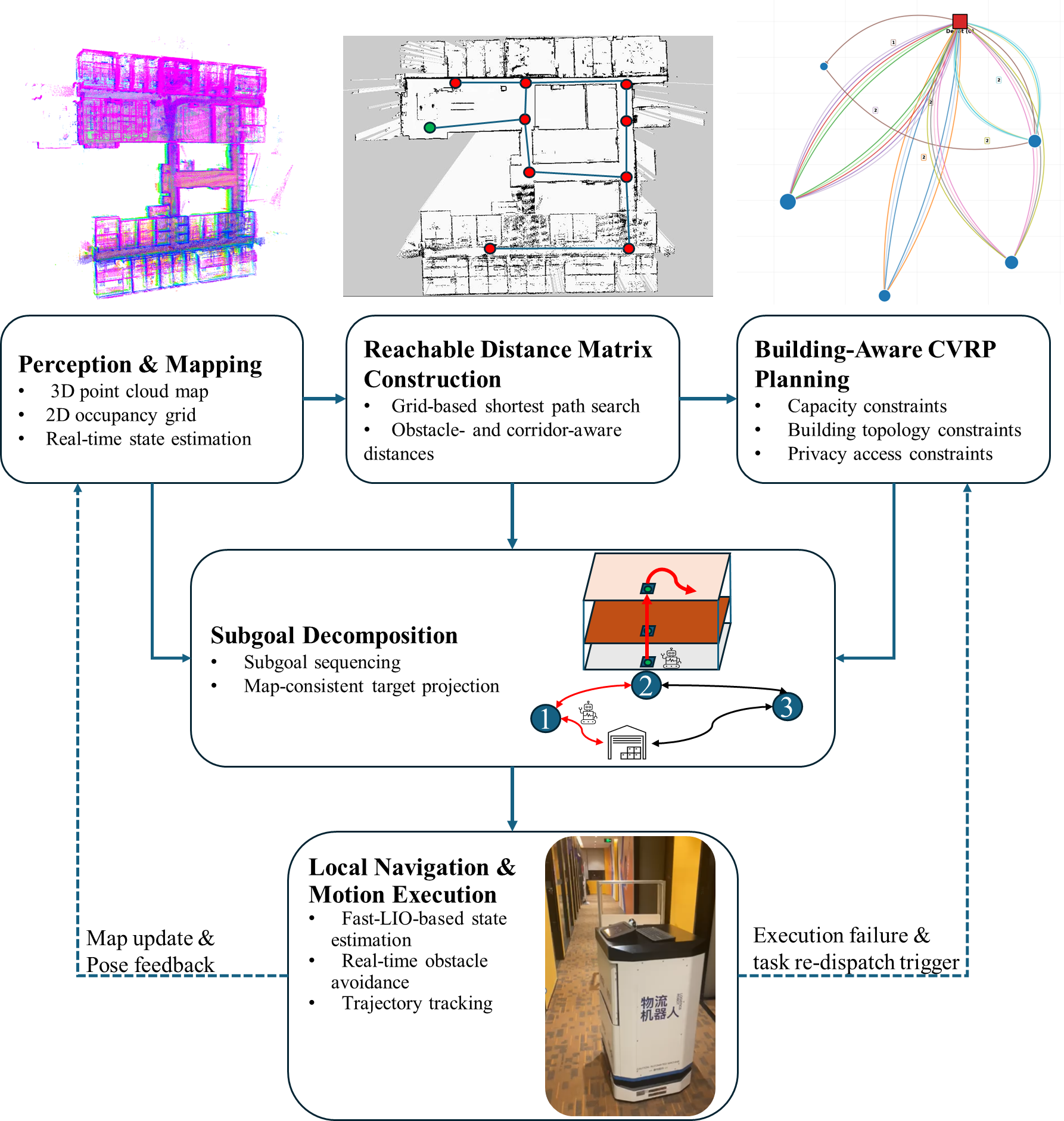}
    \caption{Hierarchical delivery framework. Modules include perception and mapping, reachable distance matrix construction, building- and privacy-aware task scheduling, subgoal decomposition, and local navigation with feedback.}
    \label{fig:overview}
\end{figure}

We propose a hierarchical delivery framework for multi-building community delivery, integrating perception, planning, and execution. The framework consists of five modules (Figure~\ref{fig:overview}): (i) \textbf{Perception and Mapping}, constructing 3D point clouds and 2D occupancy grids with real-time state estimation; (ii) \textbf{Reachable Distance Matrix Construction}, generating environment-aware distance matrices; (iii) \textbf{Building-Aware Global Task Scheduling}, allocating tasks under robot capacity $Q$ and privacy constraints; (iv) \textbf{Subgoal Decomposition}, translating global tasks into local navigation subgoals; and (v) \textbf{Local Navigation and Motion Execution}, performing closed-loop planning with feedback for map refinement and dynamic re-dispatch.

\subsection{Hierarchical Building Representation}

\begin{figure} \centering \includegraphics[width=0.5\linewidth]{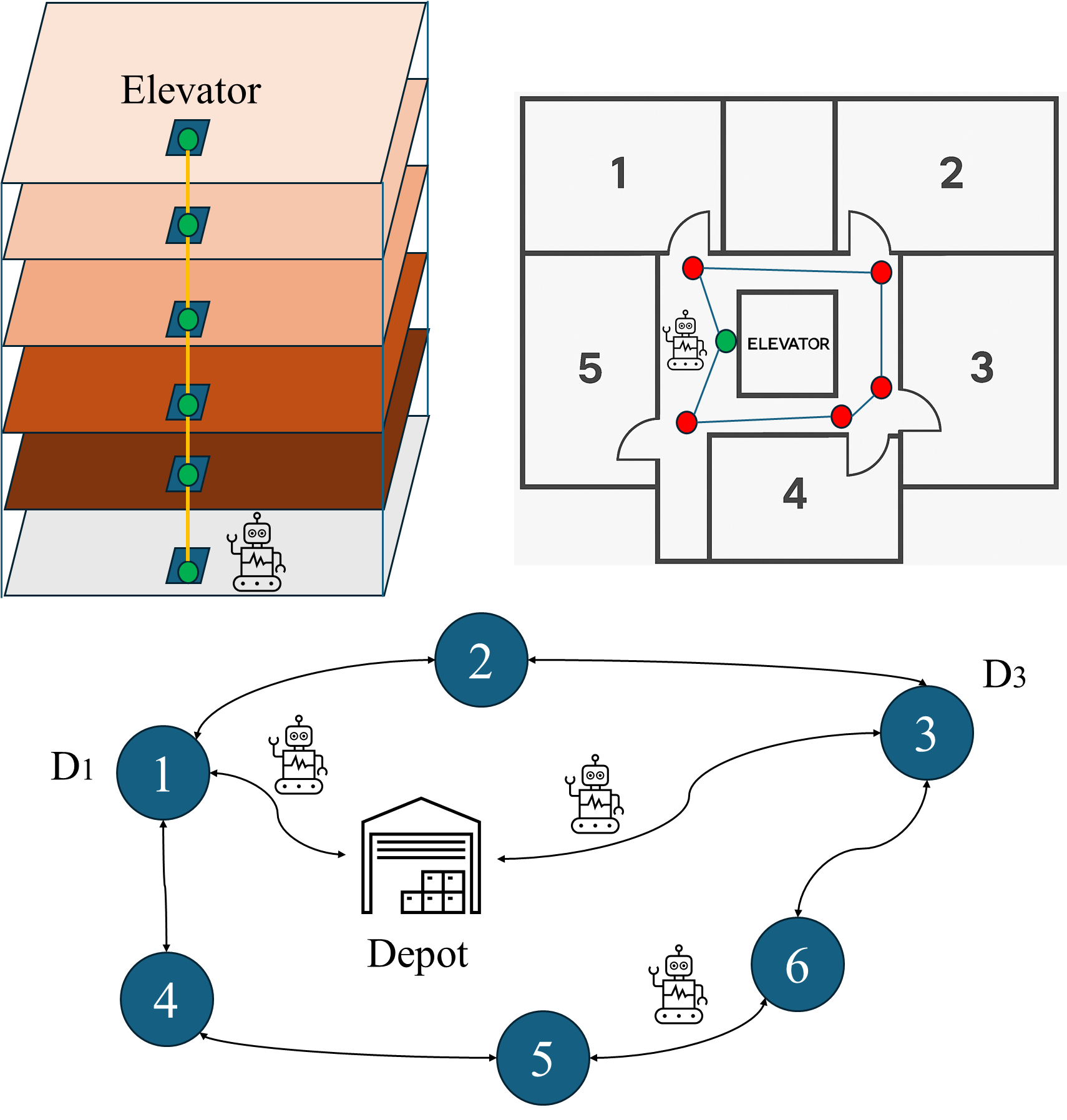} \caption{Hierarchical Building Representation} \label{fig:CVRP_3LEVEL} \end{figure}

The delivery environment is represented as a three-level hierarchical graph (shown in Fig.\ref{fig:CVRP_3LEVEL}:

\begin{itemize}
    \item \textbf{Building-Level Graph:} 
    \begin{equation}
    \label{eq:building_graph}
    G_B = (V_B, E_B),
    \end{equation}
    where $V_B$ includes the depot and building nodes, $E_B$ represents inter-building arcs, and edge cost $w_{ij}^B$ is the SLAM-derived reachable distance.

    \item \textbf{Floor-Level Graph for building $i$:}
    \begin{equation}
    \label{eq:floor_graph}
    G_F^i = (V_F^i, E_F^i),
    \end{equation}
    with nodes $V_F^i$ as floors and edges $E_F^i$ as elevator transitions. Vertical movement cost is
    \begin{equation}
    \label{eq:floor_cost}
    w_{pq}^{F,i} = \alpha |p-q| + t_\mathrm{wait}.
    \end{equation}

    \item \textbf{Room-Level Graph for floor $j$ of building $i$:}
    \begin{equation}
    \label{eq:room_graph}
    G_R^{i,j} = (V_R^{i,j}, E_R^{i,j}),
    \end{equation}
    where nodes $V_R^{i,j}$ include corridor intersections, entrances, and task locations, edges $E_R^{i,j}$ encode navigable corridors, and edge cost $w_{uv}^{R,i,j}$ is computed via shortest-path planning. Privacy-sensitive areas are penalized:
    \begin{equation}
    \label{eq:privacy_cost}
    \tilde{w}_{uv}^{R,i,j} = \lambda w_{uv}^{R,i,j}, \quad \lambda > 1.
    \end{equation}
\end{itemize}

The cumulative system-level cost is
\begin{equation}
\label{eq:total_cost}
J = \sum_{(i,j)\in E_B} w_{ij}^B + \sum_i \sum_{(p,q)\in E_F^i} w_{pq}^{F,i} + \sum_{i,j} \sum_{(u,v)\in E_R^{i,j}} \tilde{w}_{uv}^{R,i,j}.
\end{equation}

In real-world delivery scenarios, the cost structure typically satisfies the hierarchical constraint
\begin{equation}
w_{ij}^B \gg w_{pq}^{F,i} \gg \tilde{w}_{uv}^{R,i,j},
\end{equation}
which indicates that inter-building travel dominates the overall cost. This hierarchy reflects the physical scale and operational effort associated with different spatial levels and motivates the proposed hierarchical decomposition.

\subsection{Hierarchical Two-Stage Capacitated Vehicle Routing Formulation}

Leveraging the hierarchical building representation introduced previously, we formulate a \emph{building-aware} two-stage capacitated vehicle routing problem (CVRP) to address system-level delivery objectives. Rather than solving a single  routing problem across all spatial scales, we decompose the decision-making process into two distinct stages: inter-building allocation and intra-building route optimization. This hierarchical decomposition enables a top-down assignment of delivery tasks while allowing bottom-up refinement of route costs, ensuring that high-level scheduling is consistent with detailed in-building navigation.

\subsubsection{Preprocessing and Data Transformation}

Let the original delivery environment be defined as follows:
\begin{itemize}
    \item $\mathbb{A}$: the set of building-level physical arcs, corresponding to $E_B$ in Eq.~\eqref{eq:building_graph}.
    \item $\mathbb{B}$: the set of depot and building nodes, where $\mathbb{B}_0$ denotes the depot, corresponding to $V_B$ in Eq.~\eqref{eq:building_graph}.
    \item $\mathbb{C}$: the distance matrix over $\mathbb{B}$, where each entry corresponds to the building-level edge cost $w_{ij}^B$.
    \item $\mathbb{D}$: the demand associated with each node in $\mathbb{B}$, with $\mathbb{D}_0 = 0$.
    \item $\mathbbm{a}^k, \mathbbm{b}^k, \mathbbm{c}^k, \mathbbm{d}^k$: the sets of physical arcs, nodes, arc costs, and node demands within building $k \in \mathbb{B}$, respectively. The corresponding floor-level graph in Eq.~\eqref{eq:floor_graph} and room-level graph in Eq.~\eqref{eq:room_graph} are integrated in Stage~2. Specifically, $\mathbbm{c}^k$ represents the vertical movement cost $w_{pq}^{F,i}$ at the floor level and the privacy-aware cost $\tilde{w}_{uv}^{R,i,j}$ at the room level.
    \item $Q$: the capacity of each robot.
\end{itemize}

For CVRP modeling, these data are transformed as follows:
\begin{itemize}
    \item $\mathbb{A}, \mathbb{B}, \mathbb{C}, \mathbb{D} \rightarrow B, C, D$: buildings are split and duplicated according to robot capacity $Q$, producing nodes suitable for CVRP. For instance, if a building $i$ has demand $15$ and $Q=4$, it is split into four nodes with demands $[4,4,4,3]$.
    \item $\mathbbm{a}^k, \mathbbm{b}^k, \mathbbm{c}^k, \mathbbm{d}^k \rightarrow b^k, c^k, d^k$: in-building networks are transformed into complete graphs over demand nodes.  Nodes are split and replicated according to $Q$ for CVRP.
\end{itemize}

For clarity, $B, C, D$ denote inter-building CVRP data, while $b^k, c^k, d^k$ denote in-building CVRP data.

\subsubsection{Assumptions}

\begin{itemize}
    \item Each duplicated customer node is visited at most once; demand is indivisible.
    \item Robots are homogeneous with sufficient daily range; no explicit mileage constraint is imposed.
    \item Robot capacity is $Q$; daily parcel weight constraints are neglected.
    \item The number of robots is sufficient to guarantee feasible assignment.    
\end{itemize}

Above assumptions 1-3 are widely adopted assumptions for CVRP models in logistics operations research \citep{archetti2008split}. Assumption 4 can be addressed by assigning multiple trips to limited robots in real-world practice.

\subsubsection{CVRP Formulation}

\paragraph{Decision Variables:}
\begin{itemize}
    \item $x_{ij} \in \{0,1\}$: 1 if inter-building arc $(i,j)$ is traversed, 0 otherwise.
    \item $y_{ij}^k \in \{0,1\}$: 1 if in-building arc $(i,j)$ in building $k$ is traversed, 0 otherwise.
\end{itemize}

\paragraph{Stage 1: Inter-Building CVRP}  
Given duplicated building nodes $B$, arcs $C$, and demands $D_i$, the inter-building CVRP minimizes total travel cost:

\begin{equation}
\min \sum_{(i,j)\in C} C_{ij} x_{ij}
\label{eq:stage1_cost}
\end{equation}

subject to flow conservation and capacity constraints:
subject to \begin{equation} \sum_{j \in B,\, i \neq j} x_{ij} = 1, \quad \forall i \in B \setminus \{0\} \label{eq:flow_out} \end{equation} \begin{equation} \sum_{i \in B,\, i \neq j} x_{ij} = 1, \quad \forall j \in B \setminus \{0\} \label{eq:flow_in} \end{equation} \begin{equation} \sum_{j \in B \setminus \{0\}} x_{0j} = \sum_{i \in B \setminus \{0\}} x_{i0} \label{eq:depot_balance} \end{equation} \begin{equation} \sum_{j \in B \setminus \{0\}} x_{0j} = \left\lceil \sum_{i \in B} \frac{D_i}{Q} \right\rceil \label{eq:min_robots} \end{equation} \begin{equation} \sum_{i \in B \setminus S,\; j \in S,\; i \neq j} x_{ij} \ge \left\lceil \sum_{i \in S} \frac{D_i}{Q} \right\rceil, \quad \forall S \subset B \setminus \{0\} \label{eq:subtour_elim} \end{equation}
Constraints Eq. \eqref{eq:flow_out} and  \eqref{eq:flow_in} ensure that each building node has exactly one outgoing and one incoming arc. Constraint Eq. \eqref{eq:depot_balance} enforces flow balance at the depot. Constraint Eq. \eqref{eq:min_robots} specifies the minimum number of robots required to serve all demands. Constraint Eq. \eqref{eq:subtour_elim} is a generalized subtour elimination constraint (GSEC), which eliminates subtours while enforcing capacity feasibility. This constraint can be implemented as a lazy constraint to accelerate solving.

\paragraph{Stage 2: In-Building CVRP}  
For each building $k$ assigned in Stage 1, an independent in-building CVRP is solved to optimize local delivery routes:

\begin{equation}
\min \sum_{(i,j)\in c^k} c_{ij}^k y_{ij}^k
\end{equation}

subject to \begin{equation} \sum_{j \in b^k,\, i \neq j} y_{ij}^k = 1, \quad \forall i \in b^k \setminus \{0\} \label{eq:inbuilding_flow_out} \end{equation} \begin{equation} \sum_{i \in b^k,\, i \neq j} y_{ij}^k = 1, \quad \forall j \in b^k \setminus \{0\} \label{eq:inbuilding_flow_in} \end{equation} \begin{equation} \sum_{j \in b^k \setminus \{0\}} y_{0j}^k = \sum_{i \in b^k \setminus \{0\}} y_{i0}^k \label{eq:inbuilding_depot_balance} \end{equation} \begin{equation} \sum_{j \in b^k \setminus \{0\}} y_{0j}^k = \left\lceil \sum_{i \in b^k} \frac{d_i^k}{Q} \right\rceil \label{eq:inbuilding_min_robots} \end{equation} \begin{equation} \sum_{i \in b^k \setminus S,\; j \in S,\; i \neq j} y_{ij}^k \ge \left\lceil \sum_{i \in S} \frac{d_i^k}{Q} \right\rceil, \quad \forall S \subset b^k \label{eq:inbuilding_subtour_elim} \end{equation} \begin{equation} \sum_{(i,j)\in a} y_{ij}^k + \sum_{(i,j)\in a'} y_{ij}^k \le |a| + |a'| - 1 \label{eq:inbuilding_nonfull_tours} \end{equation}

Here, Eqs.~\eqref{eq:inbuilding_flow_out}--\eqref{eq:inbuilding_subtour_elim} serve the same functions as Eqs.~\eqref{eq:flow_out}--\eqref{eq:subtour_elim} above. The last constraint  Eq. \eqref{eq:inbuilding_nonfull_tours} prevents multiple non-full-capacity tours within the same building, where $a$ and $a'$ are any two different non-full capacity tour arcs’ sets during the solving process, maintaining consistency with the demand duplication preprocessing.

\subsubsection{Validity and Complexity}
\label{sub.Complexity}

Sequential optimization is efficient because we separate inter-building and in-building routing decisions. Let $|B| = n$ denote the number of duplicated building nodes, and let $|b^i| = m_i$ denote the number of in-building demand nodes in building $i$. The CVRP is NP-hard, and general ILP solvers rely on branch-and-bound or similar exponential-time techniques, whose worst-case search space grows rapidly with problem size and is influenced by capacity constraints and vehicle routing decisions.

While the exact worst-case complexity cannot be expressed in closed form, the hierarchical decomposition substantially reduces the effective problem size. Specifically, the first-stage ILP has $n^2$ binary variables, and each second-stage ILP for building $i$ has $m_i^2$ binary variables. By solving inter-building and in-building routes separately, the decomposition allows efficient optimization for large-scale delivery tasks.

\subsubsection{Algorithmic Procedure}

The overall two-stage CVRP procedure is summarized in Algorithm~\ref{alg:two_stage_cvrp}. The algorithm preprocesses demands, solves the inter-building CVRP, then independently solves in-building CVRPs, and finally maps solutions to the original network.

\begin{algorithm}[t]
\caption{Hierarchical Two-Stage CVRP for Delivery Robots}
\label{alg:two_stage_cvrp}
\begin{algorithmic}[1]
\State \textbf{Input:} Depot $0$, duplicated building nodes $B$, inter-building arcs $C$, building demands $D_i$, robot capacity $Q$, in-building nodes $b^k$, arcs $c^k$, in-building demands $d_i^k$
\State \textbf{Output:} Total delivery cost $J$, delivery routes $R$
\State Preprocess: split nodes according to $Q$
\State Initialize $J \leftarrow 0$, $R \leftarrow \emptyset$
\State Solve inter-building ILP $\rightarrow J_1, R_1$, update $J \leftarrow J + J_1$, $R \leftarrow R \cup R_1$
\For{each building $k$ in $R_1$}
    \State Solve in-building ILP $\rightarrow J_k, R_k$, map results to original network
    \State Update $J \leftarrow J + J_k$, $R \leftarrow R \cup R_k$
\EndFor
\State \Return $J, R$
\end{algorithmic}
\end{algorithm}

\subsection{Subgoal Decomposition}

After solving the two-stage CVRP, a set of global routes ${R}=\{R_i\}$
is obtained, which specifies the set of delivery locations but does not determine
the execution order of individual delivery demands.
We therefore introduce a subgoal decomposition stage to allocate and schedule
delivery demands along each route among multiple robots based on demand priority.

Each delivery demand $d$ is associated with a priority weight $\omega_d$, where
larger weights indicate earlier dispatch.
The priority weight captures batching efficiency, delivery urgency, and time
criticality.
In practice, multiple parcels destined for the same room are consolidated into a
single storage bin occupying one unit of robot capacity.
Let $n_d$ denote the number of parcels in demand $d$, with larger $n_d$ yielding
higher priority.
Expedited demands and demands approaching their deadlines are assigned elevated
weights.

The priority weight is defined as
\begin{equation}
\omega_d
=
\alpha n_d
+
\beta \mathbb{I}_d^{\mathrm{exp}}
+
\gamma f(t_d),
\end{equation}
where $\mathbb{I}_d^{\mathrm{exp}}$ indicates expedited demands and $f(t_d)$ is a
monotonically increasing function of delivery urgency.

For each route $R_i \in {R}$, a route-level priority is computed as
\begin{equation}
\Omega(R_i) = \sum_{d \in R_i} \omega_d.
\end{equation}

Let $\mathcal{K}$ denote the set of robots.
Routes are assigned to robots in descending order of route-level priority
$\Omega(R_i)$.
Each route $R_i$ is assigned to the robot with the minimum current workload,
subject to capacity constraints.
Formally, the assignment is defined by a mapping
$\mathcal{A}:\mathcal{R}\rightarrow\mathcal{K}$.
Within each assigned route, delivery demands are executed in descending order of
$\omega_d$.

\section{Unified Mapping and Autonomous Navigation System}
\label{sec:unified}

To realize the planned delivery routes and execute the allocated demands, this section presents a unified perception--mapping--planning--control pipeline for reliable autonomous navigation in complex multi-building, multi-floor indoor environments. 

%====================================================================
\subsection{Hardware Design}
\label{subsec:hardware}

\begin{figure}[t]
    \centering
    \includegraphics[width=1\linewidth]{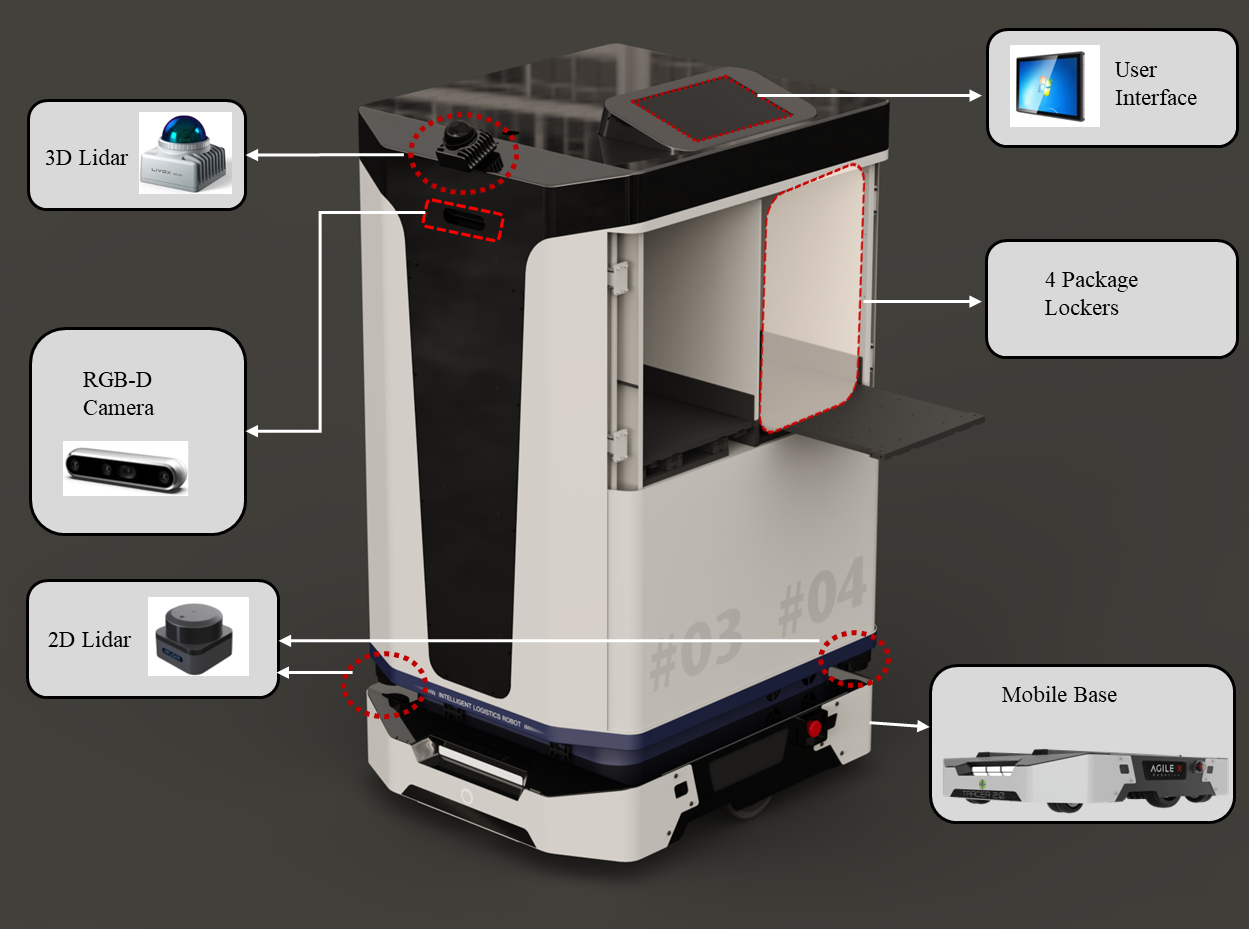}
    \caption{Hardware configuration of the autonomous delivery robot.}
    \label{fig:hardware}
\end{figure}

The hardware configuration of the autonomous delivery robot is shown in Fig.~\ref{fig:hardware}. The platform is designed for reliable operation in indoor multi-building and multi-floor environments, and integrates perception, interaction, storage, and mobility modules into a compact and unified system.

The robot is equipped with a heterogeneous sensor suite, including one 3D LiDAR, two 2D LiDARs, and an RGB camera, providing complementary sensing capabilities for environment perception, localization, and obstacle detection. This multi-modal configuration enhances robustness under varying lighting conditions and in cluttered indoor spaces. A user interface mounted on the upper body supports task interaction, system status monitoring, and manual intervention when necessary.

For parcel transportation, the robot incorporates four independent \emph{package lockers}, enabling concurrent storage and delivery of up to four parcels. The lockers are rigidly integrated into the robot body to ensure mechanical stability during navigation and vertical transitions, such as elevator usage.

All sensing, interaction, and storage modules are mounted on a differential-drive mobile base, which provides efficient and maneuverable locomotion in confined indoor environments, including corridors, elevators, and narrow passageways, while maintaining operational safety.

\begin{figure}[t]
    \centering
    \includegraphics[width=1\linewidth]{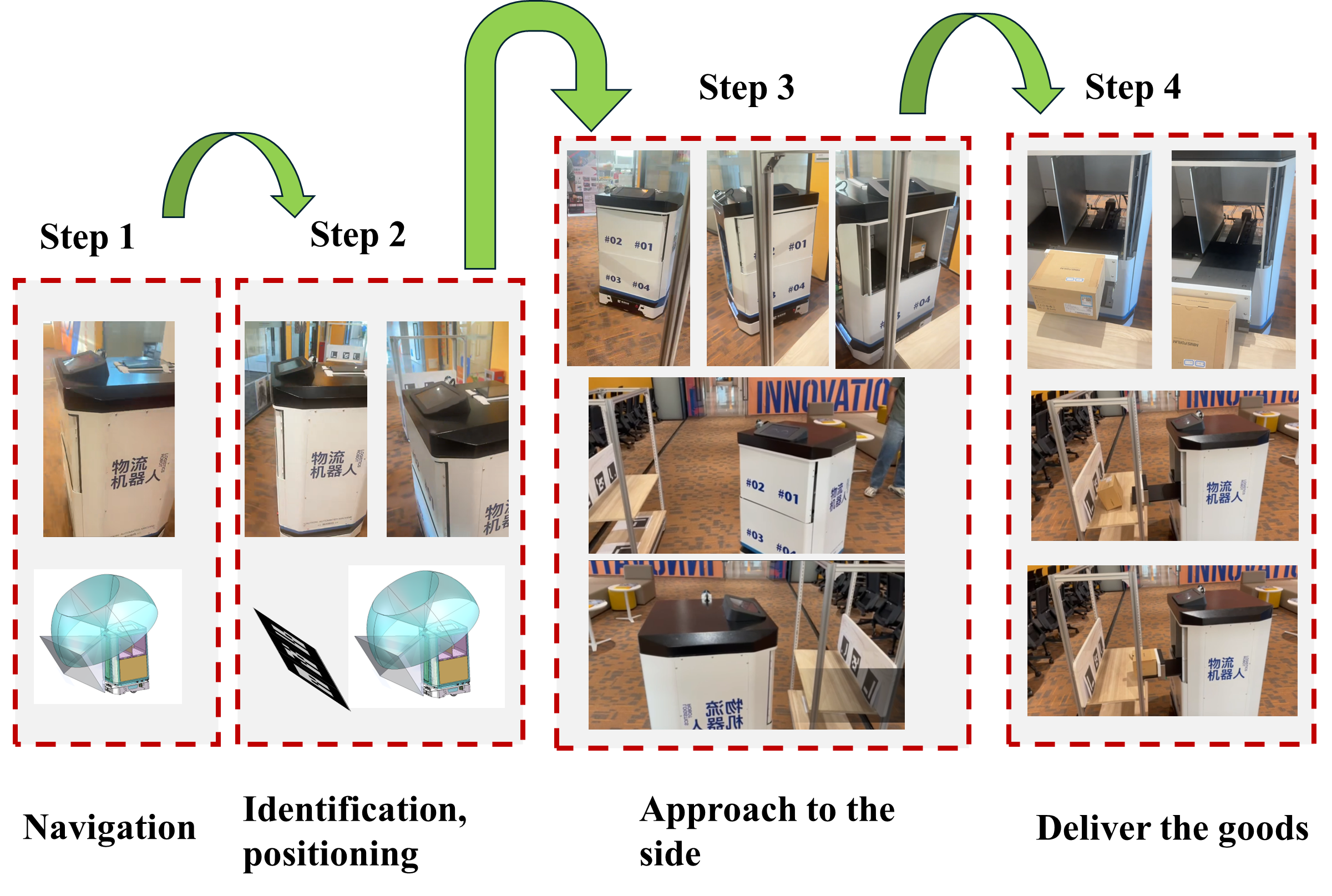}
    \caption{Parcel auto delivery mechanism at the final stage.}
    \label{fig:edgeapproch}
\end{figure}

Fig.~\ref{fig:edgeapproch} illustrates the delivery mechanism used in the final stage of each mission. Upon arriving near the target room, the robot executes an \emph{edge approach} and actuates a built-in pushing mechanism to deliver the parcel onto a desk or the floor without requiring direct user interaction. The mechanism consists of a planar push surface integrated into the locker, enabling controlled and reliable parcel release.

This contactless delivery design eliminates the need for face-to-face interaction, thereby preserving user privacy and reducing social discomfort. In addition, the robot does not need to wait for user arrival, which improves delivery efficiency and reduces overall task completion time.

The system integrates sensor calibration, tightly coupled odometry,
navigation-oriented 2D map generation, and hierarchical planning into a
coherent framework, enabling robots to traverse and service assigned subgoals
efficiently and safely.

\begin{figure}[t]
    \centering
    \includegraphics[width=1\linewidth]{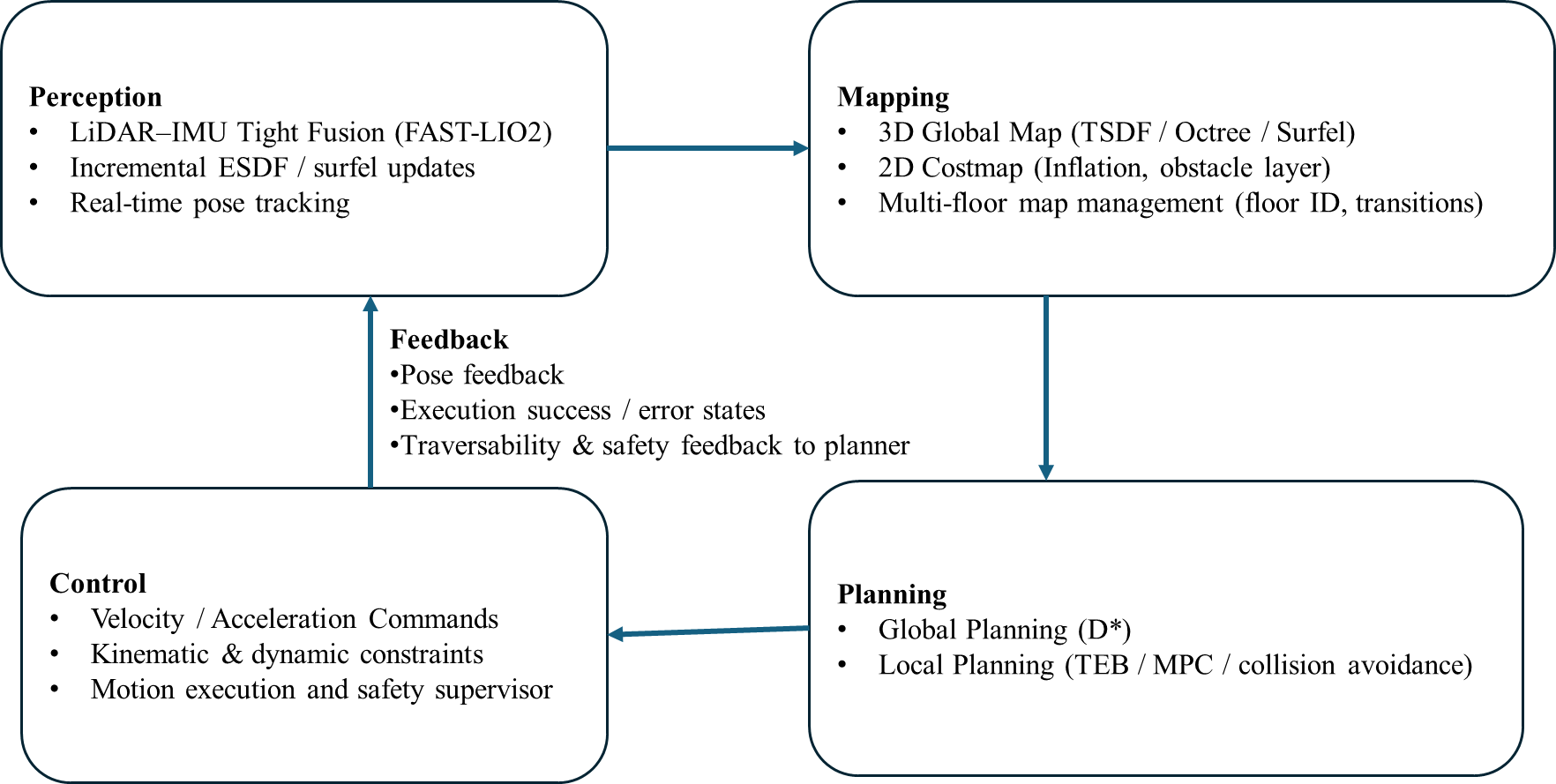}
    \caption{Overall architecture of the unified mapping and navigation system.}
    \label{fig:unifiedcontrol}
\end{figure}

%====================================================================
\subsection{Sensor Fusion and Extrinsic Calibration}
  
Robust indoor navigation relies on accurate spatial alignment among heterogeneous sensors. Precise extrinsic calibration is essential to maintain geometric consistency across perception, mapping, and motion-planning modules.

We denote the pose of sensor $S_i$ relative to the robot base frame $\mathcal{B}$ as 
$\mathbf{T}_{\mathcal{B}}^{S_i}\!\in\!\mathrm{SE}(3)$ and estimate these parameters through a joint optimization procedure.  
Given a set of time-synchronized measurements, the extrinsic calibration solves:
\begin{equation}
    \mathbf{T}_{\mathcal{B}}^{S_i^\star} = 
    \arg\min_{\mathbf{T}_{\mathcal{B}}^{S_i}}
    \sum_{k} \left\Vert 
    \mathbf{z}_{k}^{S_i}
    - 
    \hat{\mathbf{z}}_{k}\!\left(\mathbf{T}_{\mathcal{B}}^{S_i}\right)
    \right\Vert^2,
\end{equation}
where $\mathbf{z}_{k}^{S_i}$ denotes raw observations and $\hat{\mathbf{z}}_k(\cdot)$ represents their predicted geometric projections under candidate extrinsics.

The calibrated transforms $\{\mathbf{T}_{\mathcal{B}}^{S_i^\star}\}$ ensure that subsequent LiDAR points, IMU readings, and wheel odometry (if available) are expressed in a common kinematic frame, enabling consistent map construction and cross-modal fusion.

This module establishes a geometrically consistent multi-sensor frame structure, providing reliable input for downstream odometry, mapping, and navigation.

%====================================================================
\subsection{Tightly Coupled Odometry and 3D Mapping}
  
Accurate and drift-resilient state estimation is fundamental for navigation in indoor spaces, especially across floors and long corridors where geometric degeneracy frequently occurs.

The robot state is defined as: 
\begin{equation}
    \mathbf{x} = 
    \left[
    \mathbf{p}^\top,\;
    \mathbf{v}^\top,\;
    \mathbf{q}^\top,\;
    \mathbf{b}_a^\top,\;
    \mathbf{b}_g^\top
    \right]^\top,
\end{equation}
where $\mathbf{p}$, $\mathbf{v}$ are position and velocity, $\mathbf{q}$ represents attitude (unit quaternion), and $\mathbf{b}_a$, $\mathbf{b}_g$ denote IMU biases.

FAST-LIO2 provides a tightly coupled LiDAR--IMU odometry by integrating continuous IMU propagation with point-to-map residuals. The IMU prediction follows:
\begin{equation}
    \dot{\mathbf{x}} = f(\mathbf{x}, \mathbf{u}_{\text{IMU}}),
\end{equation}
and each LiDAR point $\mathbf{p}_i$ contributes a measurement residual:
\begin{equation}
    \mathbf{r}_i =
    \mathbf{p}_i^{\mathcal{W}}
    -
    \Pi_{\mathcal{M}}\!\left(
    \mathbf{T}_{\mathcal{W}}^{\mathcal{B}}
    \cdot
    \mathbf{p}_i^{\mathcal{B}}
    \right),
\end{equation}
where $\Pi_{\mathcal{M}}(\cdot)$ denotes projection onto the local geometric map $\mathcal{M}$.
State updates are solved via an iterated Kalman filter, which refine the filter's prediction using new sensor measurements, resulting in a maximum a posteriori (MAP) estimate. More details can be found in \cite{xu2022fast}.

The accumulated point clouds are registered into a dense 3D map, providing persistent structural awareness for multi-floor navigation.

This module delivers drift-minimized odometry and a locally consistent 3D map, forming the geometric backbone of navigation and global–local planning integration.

%====================================================================
\subsection{Mapping}

Although dense 3D maps preserve rich geometric structure, efficient indoor navigation and task-level planning require a compact 2D representation that supports global routing, local obstacle avoidance, and real-time control.

\begin{figure}[t]
    \centering
    \includegraphics[width=1\linewidth]{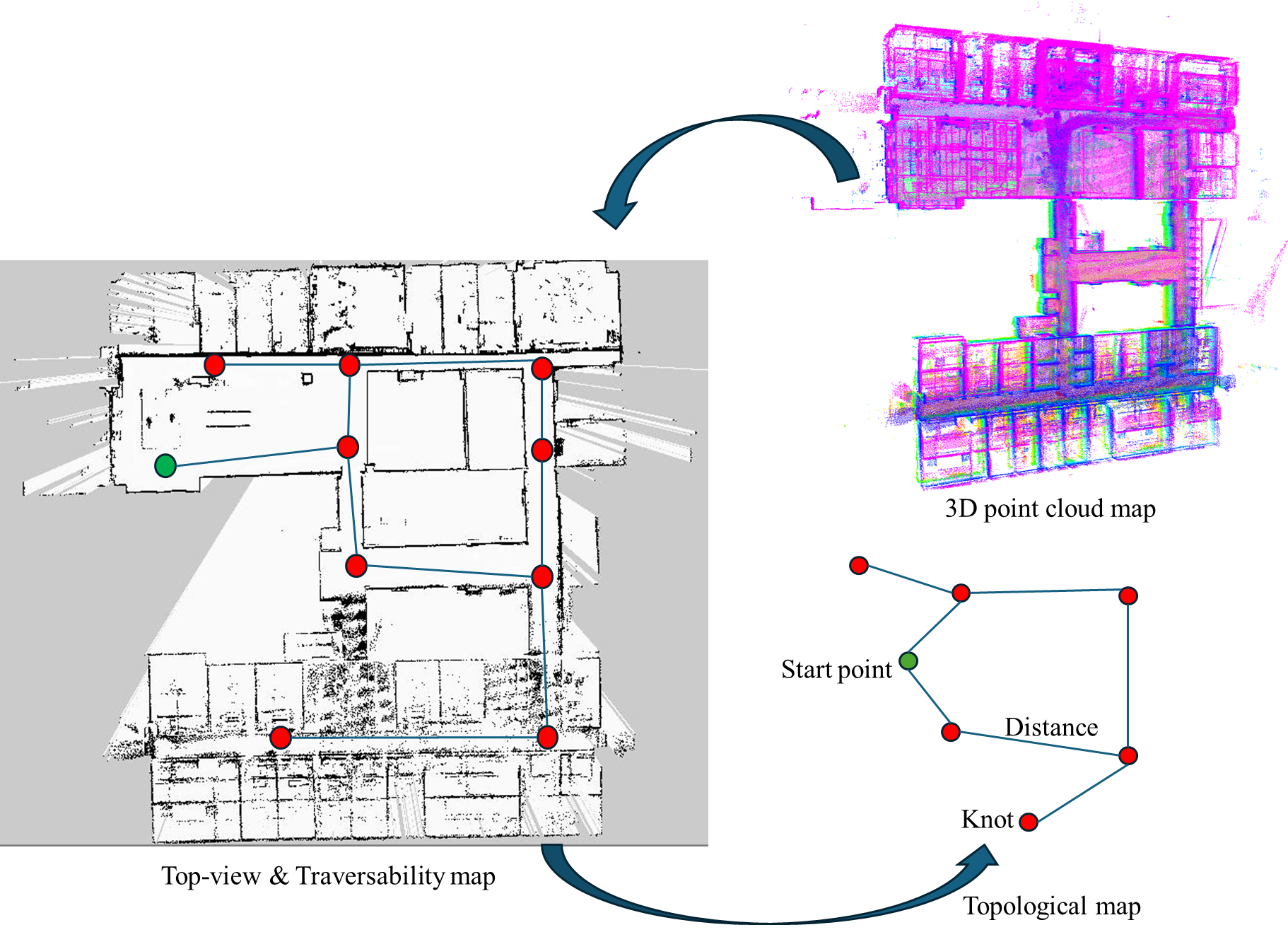}
    \caption{Mapping pipeline. Raw 3D points are projected into a top-view occupancy and cost map, which is further abstracted into a topological graph used by the CVRP-based planner.}
    \label{fig:mapconvert}
\end{figure}

As illustrated in Fig.~\ref{fig:mapconvert}, the mapping process starts from a 3D point cloud and progressively converts it into a 2D occupancy map, a traversability-aware costmap, and finally a topological representation suitable for high-level planning.

Specifically, 3D points are projected onto the horizontal plane via height filtering:
\begin{equation}
    \mathcal{P}_{2\mathrm{D}}
    =
    \big\{
    (x,y)\;|\;
    (x,y,z)\in\mathcal{P}_{3\mathrm{D}},\;
    z_{\min}\!\le z \le z_{\max}
    \big\},
\end{equation}
where $[z_{\min}, z_{\max}]$ defines the valid height range for navigable obstacles. The filtered points are discretized into a 2D grid map. For each grid cell $c$, the occupancy probability is updated using a log-odds formulation:
\begin{equation}
    L(c) = L(c)^{-} + \ell_{\text{hit/miss}},
\end{equation}
resulting in a 2D occupancy map $\mathcal{O}$ that serves as the basis for motion planning.

On top of $\mathcal{O}$, a navigation costmap is constructed by integrating multiple semantic layers:
\begin{equation}
    \mathcal{C}(c)
    = 
    \mathcal{C}_{\text{static}}(c)
    +
    \mathcal{C}_{\text{inflation}}(c)
    +
    \mathcal{C}_{\text{dynamic}}(c),
\end{equation}
where $\mathcal{C}_{\text{static}}$ encodes static obstacles derived from the occupancy map, $\mathcal{C}_{\text{dynamic}}$ captures short-term environmental changes (e.g., moving pedestrians), and $\mathcal{C}_{\text{inflation}}$ enforces safety margins around obstacles:
\begin{equation}
    \mathcal{C}_{\text{inflation}}(c)
    =
    \exp\!\left(
        -\alpha \cdot d(c)
    \right),
\end{equation}
with $d(c)$ denoting the Euclidean distance from cell $c$ to the nearest obstacle and $\alpha$ a scaling parameter controlling the decay rate.

The resulting multilayer costmap provides a compact yet expressive representation of the environment. It enables efficient global planning, safety-aware local navigation, and seamless abstraction into a topological graph for the downstream CVRP-based scheduling and routing framework.

%====================================================================
\subsection{Hierarchical Global Planning, Local Trajectory Optimization, and Execution Feedback}
  
Reliable autonomous navigation in multi-floor indoor environments requires coordinated global planning, real-time local control, and continuous feedback to ensure consistency between maps, plans, and executed motions.

\textbf{Global planning.}
Given the 2D costmap, the global planner computes a route:
\begin{equation}
    \pi^\star 
    =
    \arg\min_{\pi}
    \sum_{(u,v)\in\pi}
    C(u,v),
\end{equation}
where $C(u,v)$ is the edge traversal cost determined by $\mathcal{C}$.
Dijkstra's algorithm is utilized due to its determinism and suitability for structured indoor environments.

\textbf{Local trajectory optimization.}
At runtime, a velocity-space search (e.g., Dynamic Window Approach) evaluates feasible motion commands:
\begin{equation}
    \mathbf{u}^\star
    =
    \arg\min_{\mathbf{u}\in\mathcal{U}}
    J_{\text{tracking}}
    +
    J_{\text{obstacle}}
    +
    J_{\text{smooth}},
\end{equation}
balancing goal tracking, obstacle avoidance, and control smoothness.

\textbf{Execution Feedback and Robustness.}
Executed control commands are continuously fused into the state estimator, enabling both odometry refinement and layered costmap updates that preserve global--local consistency during navigation. To ensure robustness in challenging conditions—including elevator entrances, feature-sparse corridors, and densely populated indoor areas—the system employs adaptive feedback mechanisms such as dynamic inflation scaling, transient cost adjustments, and uncertainty-triggered global re-planning. In addition, opportunistic relocalization using visual fiducials, structural keyframes, or loop-closure cues is incorporated as EKF correction updates to suppress drift during long traversals and vertical transitions. Together, these execution, perception, and estimation feedback loops form a unified closed-loop pipeline that maintains persistent localization accuracy and stable navigation across complex multi-floor indoor environments.

This hierarchical planning and control design forms a closed-loop navigation system that adapts to dynamic environments while maintaining global consistency and safety guarantees.

\section{Experiments and Results}

This section presents a comprehensive evaluation of the proposed hierarchical autonomous indoor delivery system. Our experimental study is structured in three stages:  
(i) algorithmic verification across diverse architectural configurations,  
(ii) controlled comparative experiments isolating critical modeling components, and  
(iii) deployment in real multi-building, multi-floor environments.  
Evaluation focuses on route rationality, plan–execution consistency, hierarchical cost allocation, and robustness under practical constraints.

\subsection{Algorithm Verification}

The first stage of evaluation focuses on validating the correctness, scalability, and hierarchical consistency of the proposed CVRP-based scheduling framework. In particular, we assess whether the planner reliably generates feasible routes that respect the three-level building--floor--room cost hierarchy and correctly incorporates privacy-aware penalties into the optimization procedure.

\subsubsection{Hierarchical Solution}
\label{subsub.Solution}

% To examine whether the optimized routes conform to architectural semantics, each solution is decomposed into three interpretable motion components:

% \begin{itemize}
%     \item \textbf{Inter-building movement:} Long-range traversal across spatially separated buildings, typically associated with the highest travel cost.
%     \item \textbf{Elevator-driven vertical transitions:} Floor-to-floor movement via elevators, where both vertical offsets and waiting-time penalties are explicitly modeled.
%     \item \textbf{Intra-floor room navigation:} Local horizontal motion within individual floors, including corridor following and final approaches to delivery rooms.
% \end{itemize}

% This decomposition reveals how travel costs are distributed across different hierarchical layers and highlights which components dominate total routing expenditure.
\begin{figure}[htbp]
    \centering
    \includegraphics[width=1\linewidth]{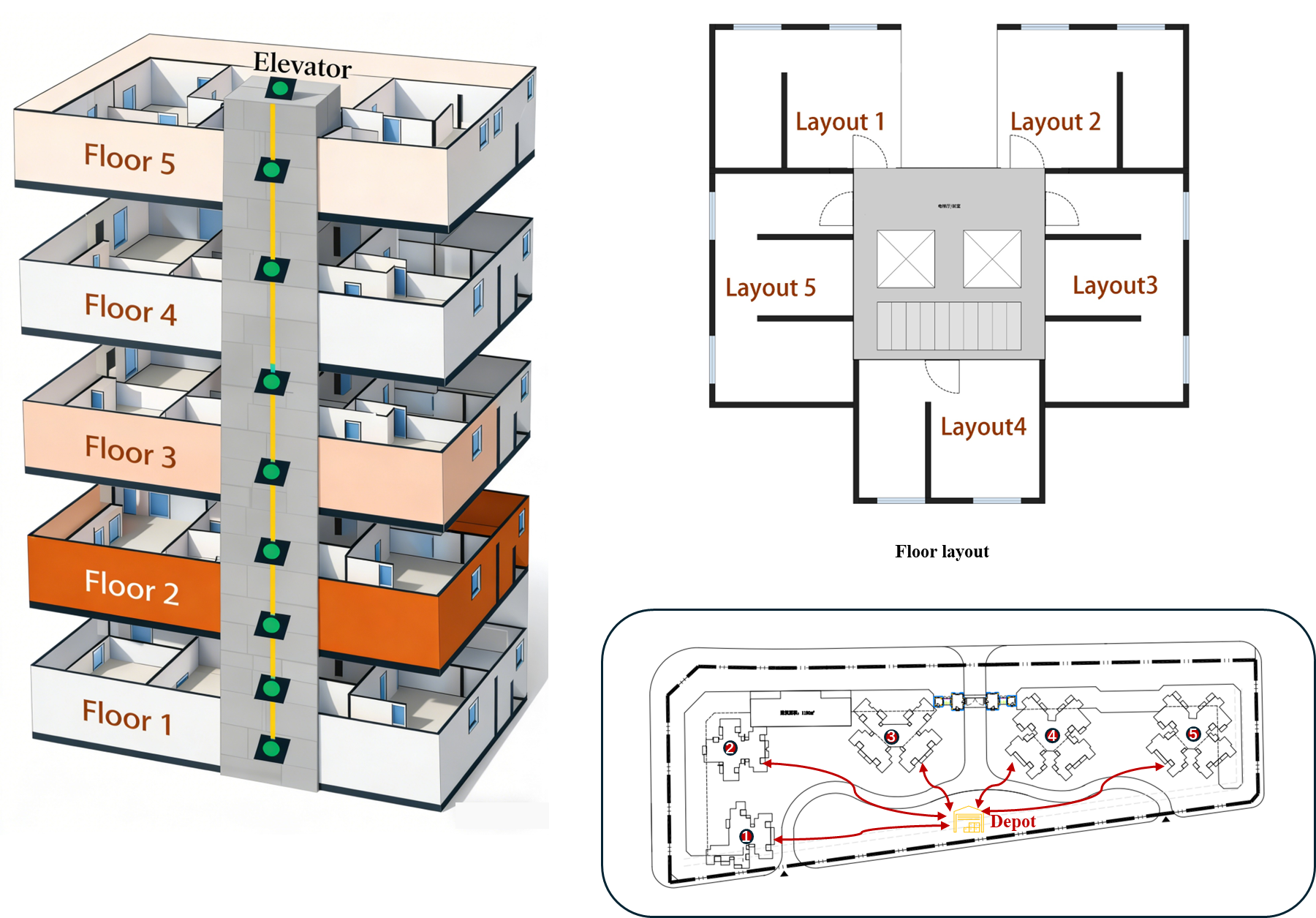}
    \caption{Hierarchical spatial representation of the residential delivery scenario.}
    \label{fig:solutionmap}
\end{figure}

Figure~\ref{fig:solutionmap} depicts the hierarchical layout of the residential community examined in this study, illustrating the spatial arrangement of five residential buildings, the floors within each building, and the internal layout of an individual building.
Although the floor-level layouts may vary across buildings in practice, all floors are assumed to share an identical configuration for simplicity. Moreover, the distance from each building to the depot is set uniformly to $100$. In Stage~1, the inter-building routing problem is formulated on an omnidirectional graph, where each building node is connected with a uniform edge weight of $100$. Stage~2 refines the solution by specifying the detailed intra-building delivery routes. Within each building, floors are connected exclusively through a single elevator, and each floor forms a closed loop consisting of five rooms. Vertical movements incur a traversal cost of $5$ per floor, along with an additional elevator waiting-time penalty of $20$, while horizontal room-to-room distances are fixed to $1$. The vehicle capacity is set to $Q = 4$, consistent with the hierarchical decomposition strategy.

\begin{figure}
    \centering
    \includegraphics[width=1\linewidth]{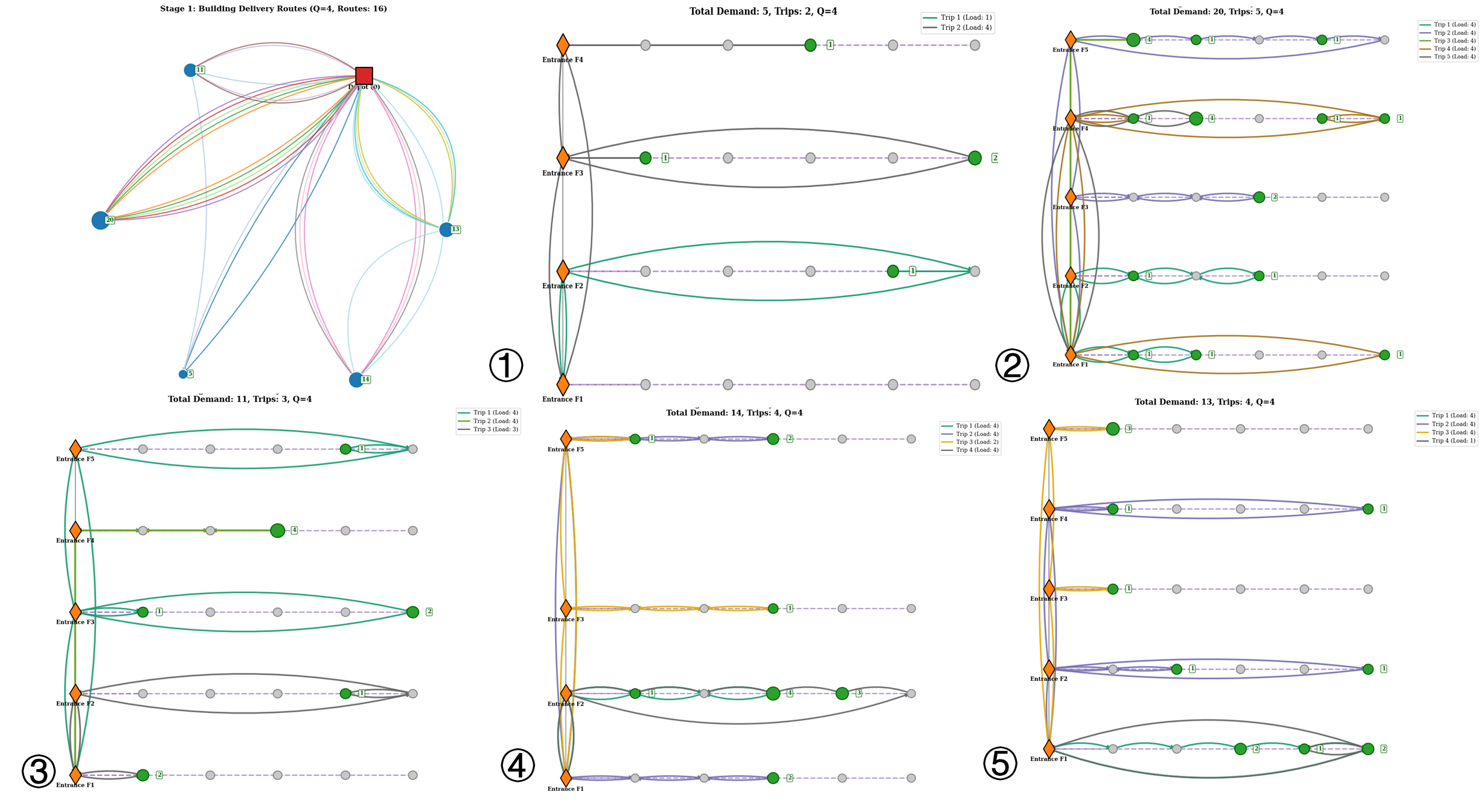}
    \caption{Hierarchical optimized delivery route.}
    \label{fig:stand}
\end{figure}

Figure~\ref{fig:stand} further presents a representative two-stage decomposition of the delivery problem over this scenario.  Under this configuration, the total cost of the Stage~1 solution is $3400$, and the corresponding routing plan is summarized in Table~\ref{tab:route_verification}.

\paragraph{Demand Decomposition and Routing Results.}

The original and decomposed demands are:
\[
\text{Demand}:\ \{0{:}\,0,\ 1{:}\,5,\ 2{:}\,20,\ 3{:}\,11,\ 4{:}\,14,\ 5{:}\,13\}.
\]

A total of sixteen building-level delivery routes are produced, as reported in Table~\ref{tab:route_verification}. For buildings whose demand is an integer multiple of the capacity \(Q\),The optimized plan assigns dedicated routes without cross-building mixing. When demand is not divisible by \(Q\), the model aggregates loads into full-capacity batches, while the remaining smaller batch forms an additional trip. This behavior reflects the intended load consolidation mechanism.

The Stage~2 (intra-building) solutions across all five buildings incur a total cost of \(1260\), with detailed routes summarized in Table~\ref{tab:building_delivery}. Notably, the Stage~2 visit orders follow the Stage~1 optimized inter-building assignments exactly, confirming the consistency of the hierarchical planner. Sub-capacity routes in Stage~2 are merged with Stage~1 routes as prescribed, demonstrating correct hierarchical alignment.

\begin{table}[h]
\centering
\caption{Building-level optimized delivery routes (capacity \(Q=4\)).}
\label{tab:route_verification}
\begin{tabular}{lcccc}
\toprule
\textbf{Route Pattern (with Demand)} & \textbf{Count} & \textbf{Total Demand} & \textbf{Distance Cost} \\
\midrule
\texttt{[0, 1 (4), 0]}                  & 1  & 4 & 200 \\
\texttt{[0, 1 (1), 3 (3), 0]}           & 1  & 4 & 300 \\
\texttt{[0, 2 (4), 0]}                  & 5  & 4 & 200 \\
\texttt{[0, 3 (4), 0]}                  & 2  & 4 & 200 \\
\texttt{[0, 4 (4), 0]}                  & 3  & 4 & 200 \\
\texttt{[0, 5 (4), 0]}                  & 3  & 4 & 200 \\
\texttt{[0, 5 (1), 4 (2), 0]}           & 1  & 3 & 300 \\
\bottomrule
\end{tabular}
\end{table}

\begin{table}[h]
\centering
\caption{Intra-building delivery routes with capacity \(Q=4\), with total demand separated into a dedicated column.}
\label{tab:building_delivery}
\begin{tabular}{
    c
    >{\raggedright\arraybackslash}p{2cm}
    >{\raggedright\arraybackslash}p{8.5cm}
    >{\raggedright\arraybackslash}p{2.3cm}
    c
}
\toprule
\textbf{Bldg} &
\textbf{Total Demand} &
\textbf{Omni-direction Routes with Demand} &
\textbf{Load per Trip} &
\textbf{Distance Cost} \\
\midrule
1 &
5 &
\texttt{[4.5 (1), 1.3 (3)]}  
\texttt{[2.4 (1)]} &
[4,1] & 132.0 \\
2 &
20 &
\texttt{[4.5 (1), 4.4 (1), 4.3 (1), 4.1 (1)]}  
\texttt{[3.2 (1), 3.3 (1), 5.1 (2)]}  
\texttt{[2.4 (1), 2.2 (2), 1.1 (1)]}  
\texttt{[1.3 (3), 1.5 (1)]}  
\texttt{[5.3 (3), 5.5 (1)]} &
[4,4,4,4,4] & 334.0 \\
3 &
11 &
\texttt{[2.5 (2), 2.4 (2)]}  
\texttt{[1.4 (2), 3.3 (1), 3.2 (1)]}  
\texttt{[5.3 (1), 5.4 (1), 4.1 (1)]} &
[4,4,3] & 232.0 \\
4 &
14 &
\texttt{[4.3 (2), 5.5 (1), 5.3 (1)]}  
\texttt{[3.4 (2), 3.5 (2)]}  
\texttt{[4.5 (1), 2.3 (2), 2.5 (1)]}  
\texttt{[1.2 (1), 1.4 (1)]} &
[4,4,4,2] & 280.0 \\
5 &
13 &
\texttt{[2.1 (1), 5.4 (1), 5.2 (2)]}  
\texttt{[2.4 (1), 4.1 (1), 4.4 (2)]}  
\texttt{[3.3 (1), 3.1 (2), 1.1 (1)]}  
\texttt{[1.3 (1)]} &
[4,4,4,1] & 282.0 \\
\bottomrule
\end{tabular}
\end{table}

% \begin{table}[h]
% \centering
% \caption{Intra-building delivery details for each building (Q=4).}
% \label{tab:building_delivery}
% \begin{tabular}{
%     p{1cm}  % 
%     p{1.5cm}  % 
%     p{3.5cm}  % Room Demand
%     p{5cm}    % Omni-direction Routes
%     p{2.2cm}  % Load per Trip
%     c
% }
% \toprule
% \textbf{Building} & \textbf{Total Demand} & \textbf{Room Demand} & \textbf{Omni-direction Routes} & \textbf{Load per Trip} & \textbf{Total Distance} \\
% \midrule
% 1 & 5 & 4.5:1, 2.4:1, 1.3:3
%   & \texttt{[1.0,4.5,1.3,1.0], [1.0,2.4,1.0]}
%   & [4,1] & 132.0 \\
% 2 & 20 & 3.3:1,4.4:1,4.5:1,3.2:1,2.4:1,1.3:3,4.1:1,4.3:1,1.5:1,5.3:3,1.1:1,5.1:2,5.5:1,2.2:2
%   & \texttt{[1.0,4.5,4.4,4.3,4.1,1.0], [1.0,3.2,3.3,5.1,1.0], [1.0,2.4,2.2,1.1,1.0], [1.0,1.3,1.5,1.0], [1.0,5.3,5.5,1.0]}
%   & [4,4,4,4,4] & 334.0 \\
% 3 & 11 & 2.4:2,3.2:1,3.3:1,2.5:2,4.1:1,1.4:2,5.3:1,5.4:1
%   & \texttt{[1.0,2.5,2.4,1.0], [1.0,1.4,3.3,3.2,1.0], [1.0,5.3,5.4,4.1,1.0]}
%   & [4,4,3] & 232.0 \\
% 4 & 14 & 4.3:2,2.3:2,1.4:1,3.4:2,4.5:1,1.2:1,2.5:1,5.3:1,3.5:2,5.5:1
%   & \texttt{[1.0,4.3,5.5,5.3,1.0], [1.0,3.4,3.5,1.0], [1.0,4.5,2.3,2.5,1.0], [1.0,1.2,1.4,1.0]}
%   & [4,4,4,2] & 280.0 \\
% 5 & 13 & 2.1:1,2.4:1,5.4:1,3.3:1,3.1:2,5.2:2,4.1:1,1.1:1,4.4:2,1.3:1
%   & \texttt{[1.0,2.1,5.4,5.2,1.0], [1.0,2.4,4.1,4.4,1.0], [1.0,3.3,3.1,1.1,1.0], [1.0,1.3,1.0]}
%   & [4,4,4,1] & 282.0 \\
% \bottomrule
% \end{tabular}
% \end{table}

\subsubsection{Test Scenarios}

To extensively evaluate generalization across architectural variations, the algorithm was tested under multiple layout configurations involving changes in inter-building spacing, vertical structure, elevator placement, and intra-floor geometric complexity.

\paragraph*{1) Varying building distributions.}

\begin{figure}
    \centering
    \includegraphics[width=\linewidth]{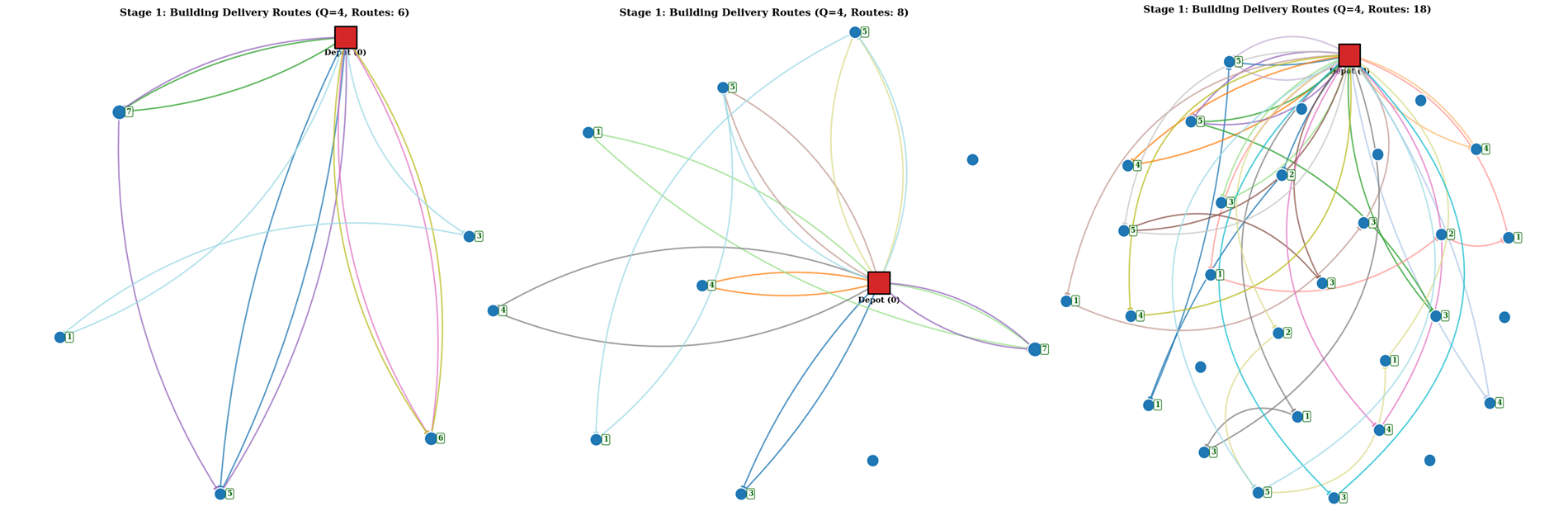}
    \caption{
    Representative variations of inter-building spatial distributions.  
    All scenarios share a single depot (red) but differ in the number and placement of buildings:  
    (a) 5 buildings, (b) 10 buildings, (c) 30 buildings.  
    Each node is annotated with its corresponding delivery demand.}
    \label{fig:df_buding}
\end{figure}

Figure~\ref{fig:df_buding} presents three representative inter-building configurations.  
The number of buildings increases from 5 to 30, which significantly alters spatial dispersion and connectivity.

\paragraph*{2) Varying multi-floor structures.}

\begin{figure}
    \centering
    \includegraphics[width=0.75\linewidth]{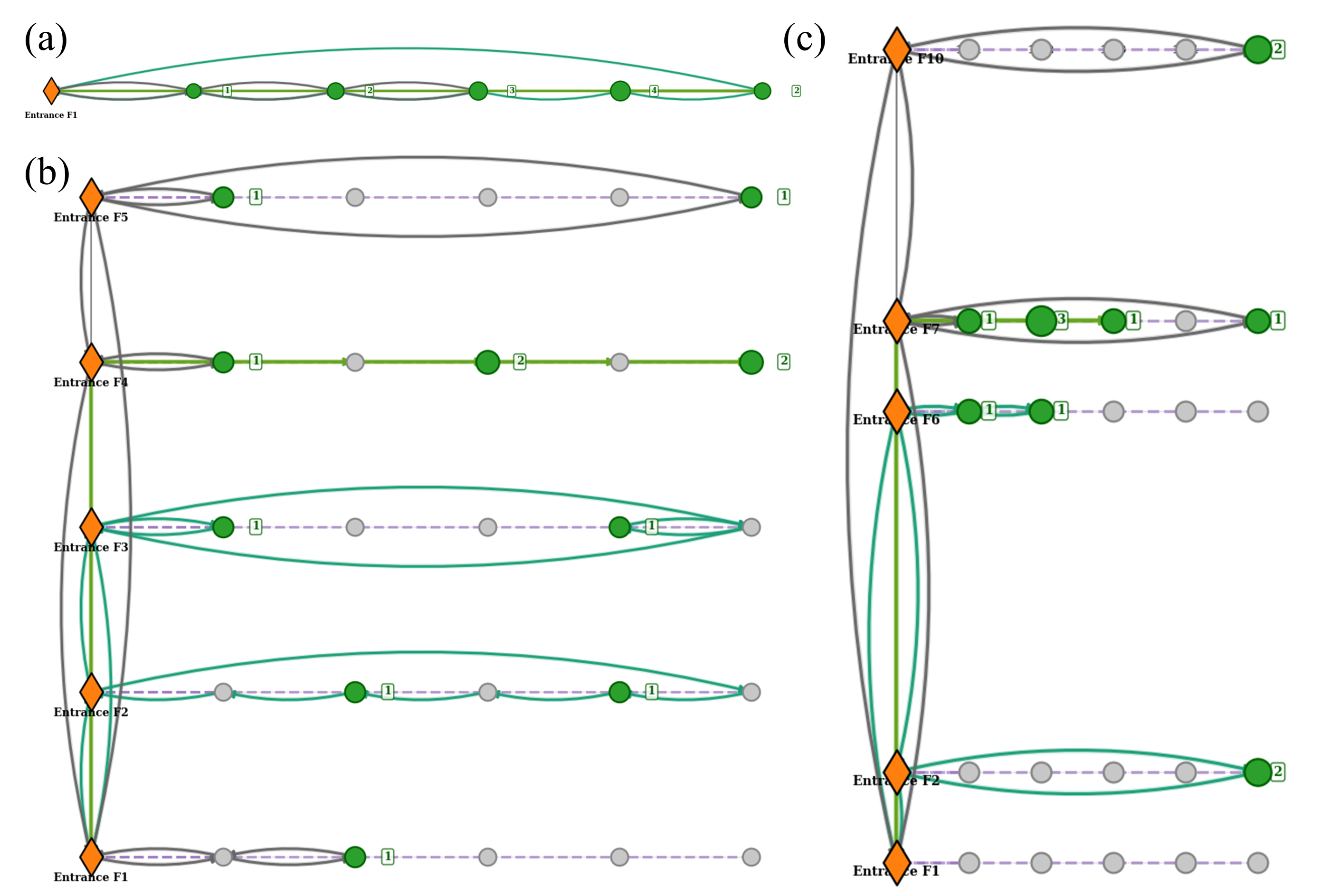}
    \caption{
    Floor-level structural variations used to evaluate the planner’s sensitivity to vertical architectural depth.  
    (a) Single floor, (b) 5 floors, and (c) 10 floors.  
    All floors share identical intra-floor layouts, with a total demand of 12 packages and capacity $Q=4$, producing three trips for all cases.}
    \label{fig:df_floor}
\end{figure}

To evaluate vertical scalability, we tested single-floor, 5-floor, and 10-floor environments (Fig.~\ref{fig:df_floor}).  
Although the number of floors changes significantly, the hierarchical planner maintains trip efficiency.

\paragraph*{3) Varying intra-floor room.}

\begin{figure}
    \centering
    \includegraphics[width=0.75\linewidth]{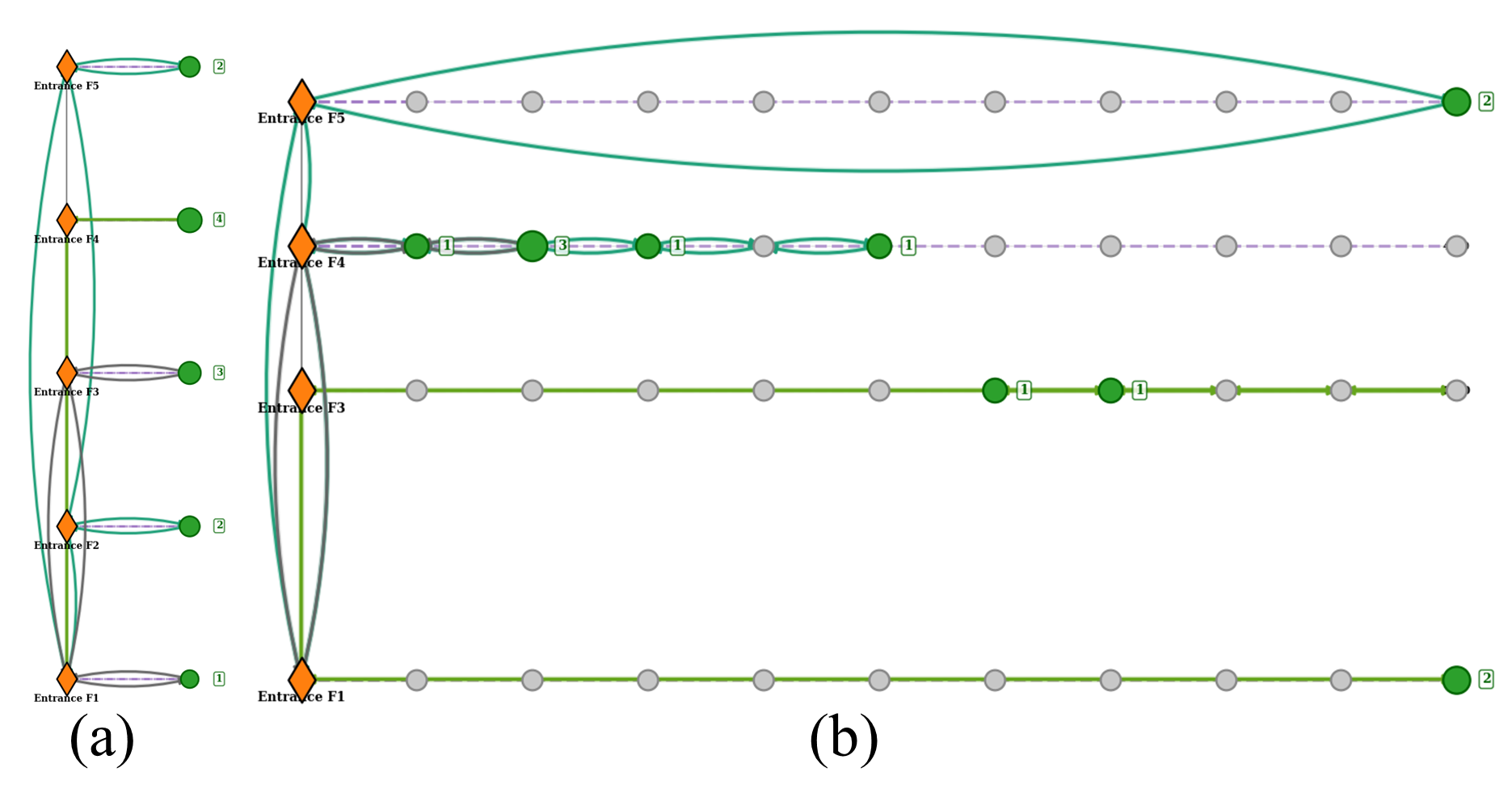}
    \caption{
    Intra-floor geometric configurations illustrating differences in room density.  
    (a) A single-room layout;  
    (b) a 10-room layout.  
    Both scenarios contain 12 total packages and capacity $Q=4$, resulting in three trips.}
    \label{fig:df_room}
\end{figure}

The planner's sensitivity to intra-floor geometric density is evaluated in Fig.~\ref{fig:df_room}.  
Despite the different number of rooms, the number of trips remains constant.

\paragraph*{4) Room-topology abstractions.}

\begin{figure}
    \centering
    \includegraphics[width=1\linewidth]{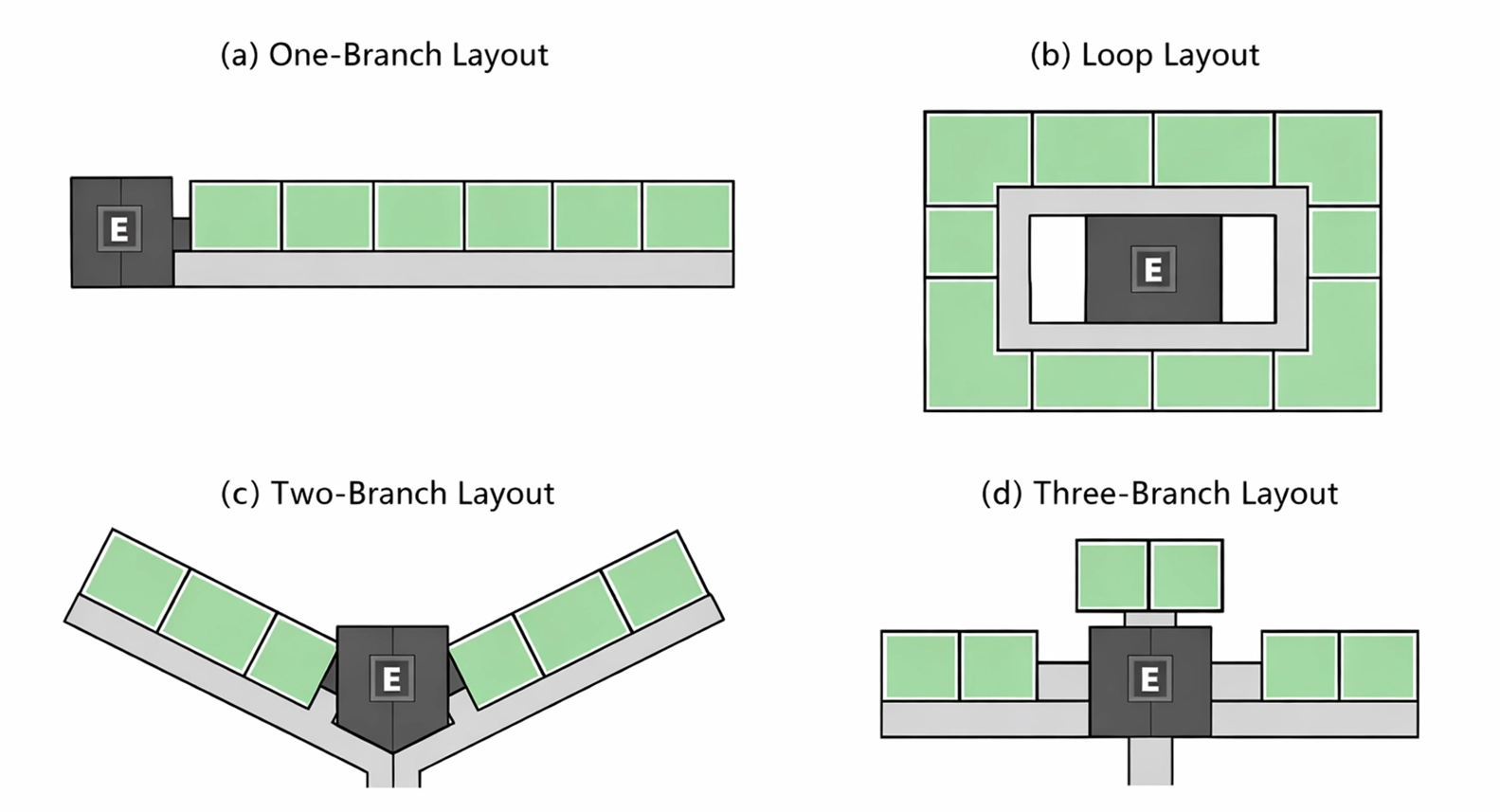}
    \caption{Classification of Residential Building Layouts Based on Branching Configurations.}
    \label{fig:roomtopology}
\end{figure}

Fig.~\ref{fig:roomtopology} presents four representative residential building layout typologies based on corridor branching patterns: (a) One-branch layout, characterized by a single linear gallery; (b) Loop layout, organized around a central core with a continuous circulation ring; (c) Two-branch (butterfly) layout, in which two symmetrical wings extend from a central hub, commonly adopted to improve daylight access and orientation; and (d) Three-branch (T-shaped) layout, featuring a tri-directional branching configuration that enables compact and efficient spatial organization.

\begin{figure}
    \centering
    \includegraphics[width=\linewidth]{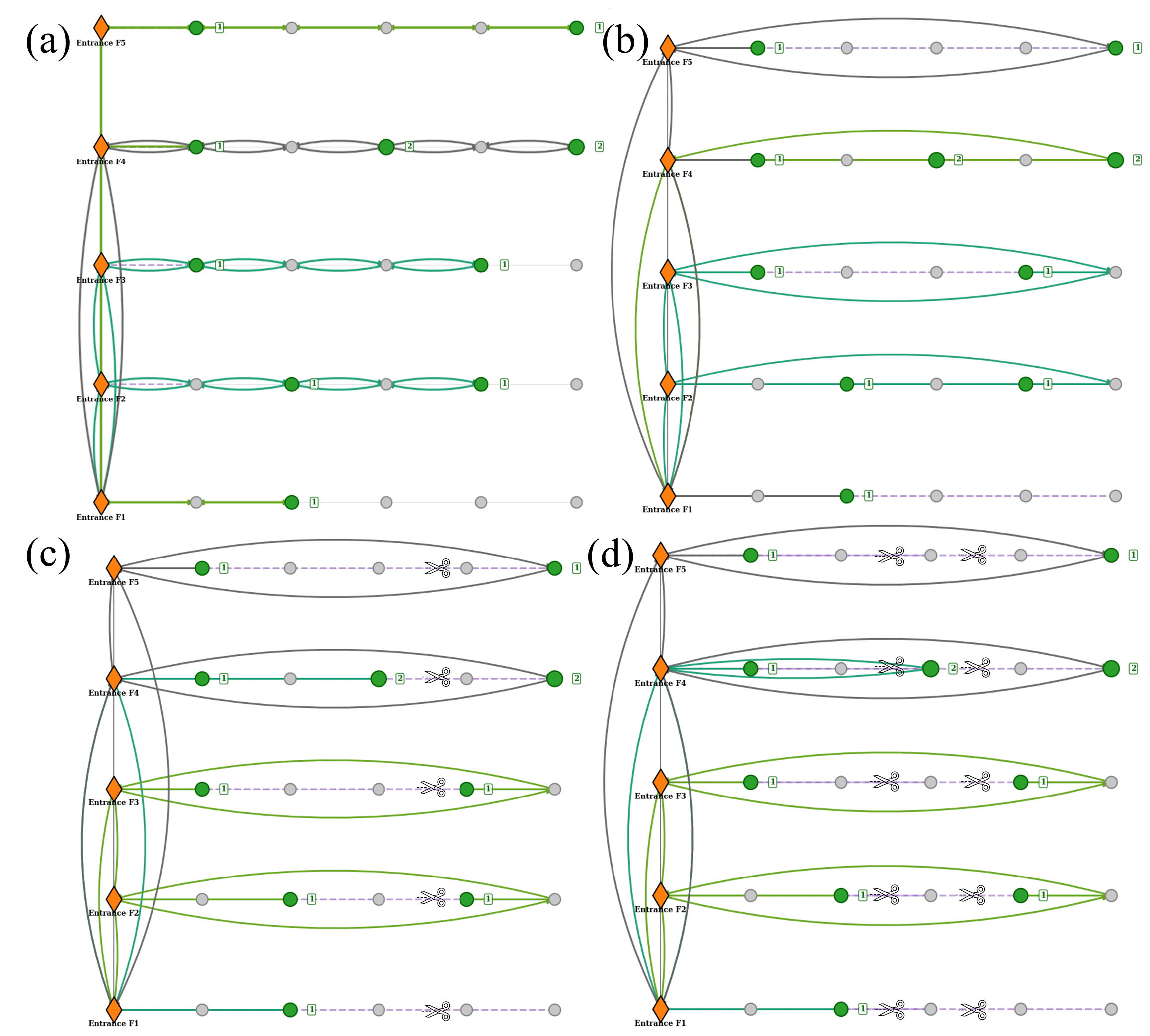}
    \caption{
    Canonical corridor-topology abstractions used to assess the structural generality of the scheduling algorithm.  
    (a) One-branch layout, (b) loop layout,  
    (c) two-branch layout (3--2 separation), and  
    (d) three-branch layout (2--1--2 separation).  
    All scenarios contain 12 packages and capacity $Q=4$, yielding three delivery trips.}
    \label{fig:df_layout}
\end{figure}

To isolate the impact of corridor topology, a set of canonical room-level graph structures was constructed, as illustrated in Fig.~\ref{fig:df_layout}. All configurations show identical vertical transfer costs, such that performance differences arise solely from room-level connectivity. The distance between adjacent rooms is normalized to one unit.

Under this setting, the room-level distance costs for layouts (a)–(d) are 42, 28, 30, and 28, respectively.
The one-branch layout yields the highest cost due to repeated backtracking along the corridor. The two-branch layout reduces the distance cost by introducing additional connectivity, while the loop and three-branch layouts achieve the lowest costs. In particular, the loop structure allows the robot to reach terminal rooms without requiring a return along the same path, and increasing the number of branches effectively shortens the average distance to the starting point. These results indicate that layouts with higher connectivity—ideally with all rooms directly connected to the elevator core—are more favorable for autonomous delivery.

Across all configurations, the planner consistently generates feasible and efficient solutions, demonstrating robustness and confirming that the hierarchical cost model is correctly implemented.

\subsection{Comparative Experiments}

We further conducted controlled studies to isolate and quantify the influence of key modeling components.

\subsubsection{Comparative Analysis of Delivery Loop Strategies}
\label{subsec:comparison}

This subsection compares different delivery loop strategies in a multi-building indoor environment consisting of a central depot and four multi-story buildings. 
Each delivery robot has a maximum carrying capacity of $Q=4$ packages, the other building graph parameters are set as section \ref{subsub.Solution}, and the total weighted travel distance is used as the evaluation metric.

The delivery task is decomposed into two stages. 
Stage~1 corresponds to inter-building transportation between the depot and building entrances, while Stage~2 focuses on intra-building delivery, including vertical (floor-to-floor) and horizontal (room-to-room) traversal.

\paragraph{Delivery Strategies}
Five representative strategies are evaluated.

\textbf{FIFO Baseline:}
Packages are dispatched in a first-in-first-out (FIFO) order. 
A single tour may include packages for multiple buildings and floors, leading to inefficient inter-building traversal and highly suboptimal intra-building routes.

\textbf{Sequential:}
Packages are delivered sequentially by building ID, and floors within each building are also served in a fixed order. 
Most tours follow the pattern $\textit{Depot} \rightarrow B_i \rightarrow \textit{Depot}$, which improves clustering at the building level but does not optimize floor- or room-level routing.

\textbf{Hybrid Strategy (Optimized Stage~1 + FIFO Stage~2):}
This strategy applies an optimized solution for Stage~1 inter-building routing to improve capacity utilization across buildings, while Stage~2 intra-building delivery remains FIFO ordered. 
As a result, inter-building travel is reduced, but inefficient intra-building routing still dominates the overall cost.

\textbf{Global CVRP:}
This strategy abandons the two-stage decomposition and formulates a single CVRP over the complete room-level network. 
All rooms of all buildings are embedded, together with the depot, in one distance matrix, and inter-building and intra-building movements are optimized simultaneously within one set of capacity-constrained tours. 
It constitutes the global counterpart of the hierarchical model and serves as a reference for assessing the loss or gain induced by decomposition.

\textbf{Hierarchical CVRP Model:}
The proposed approach formulates the delivery task as a hierarchical capacitated vehicle routing problem (CVRP). 
Inter-building routing is optimized to allow partial loads across buildings, while intra-building delivery jointly optimizes floor and room visitation sequences before returning to the building entrance. 
By optimizing both stages in a unified framework, the model achieves globally efficient delivery loops.

\begin{figure}[t]
    \centering
    \includegraphics[width=1\linewidth]{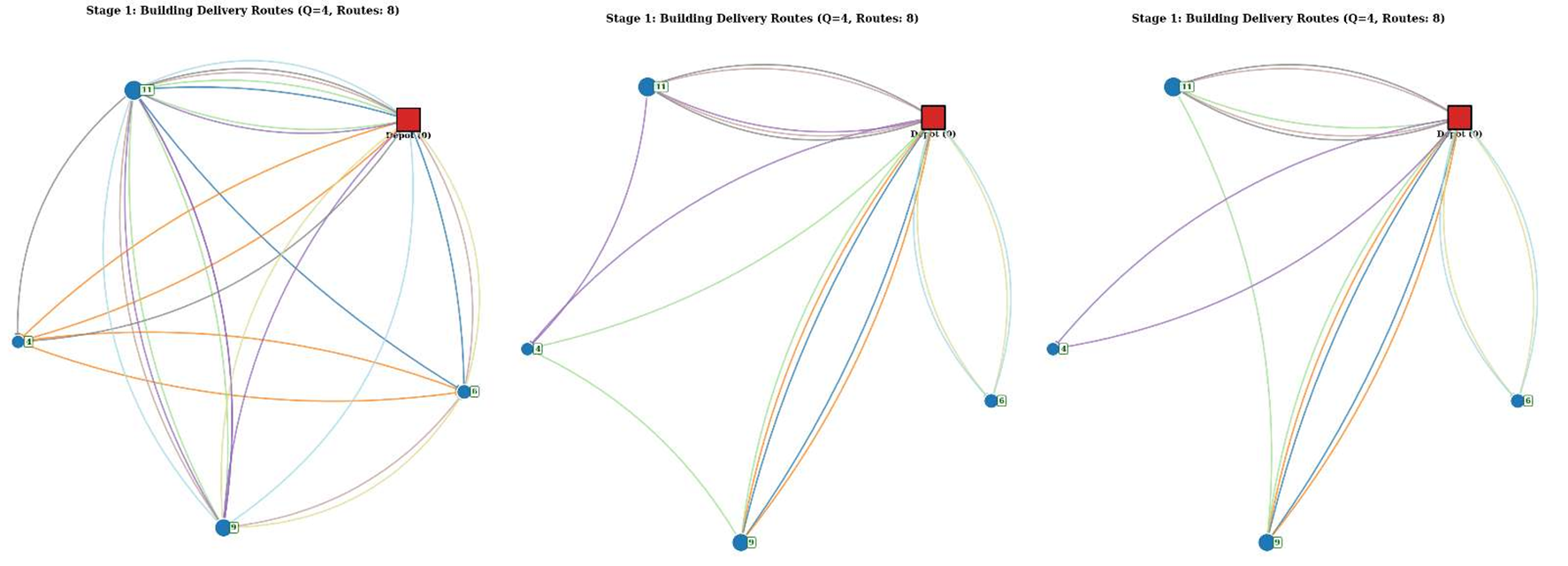}
    \caption{Inter-building delivery loops for four strategies: (a) FIFO Baseline; (b) Sequential ; (c) Global CVRP and Hierarchical CVRP Model.}
    \label{fig:comparison_models}
\end{figure}

Table~\ref{tab:comparison} summarizes the delivery costs of different strategies, with all results compared against the FIFO Baseline. Fig.~\ref{fig:comparison_models} provides a visual comparison of the corresponding inter-building routing patterns.

The FIFO Baseline yields the highest total cost due to the absence of spatial or capacity-aware organization, resulting in frequent inter-building switching and redundant intra-building traversal. It serves as a benchmark reference for delivery efficiency.

The Sequential reduces the total travel cost by $46.0\%$ relative to the FIFO Baseline. In this strategy, deliveries are strictly ordered first by building ID and further by floor and room indices within each building. This hierarchical sorting significantly improves route clustering and reduces unnecessary revisits, indicating that even simple deterministic heuristics can achieve substantial efficiency gains in structured indoor environments.

The Hybrid strategy, which applies efficient routing at the inter-building level (Stage~1) while using FIFO delivery sequences within buildings (Stage~2), achieves a $40.2\%$ cost reduction. Although its Stage~1 cost matches that of the optimized model, the lack of intra-building optimization leads to excessive corridor traversal and elevator usage. As a result, its overall performance is inferior to the Sequential strategy, highlighting that partial optimization across stages is insufficient.

The \textbf{Hierarchical CVRP Model} achieves the lowest total cost, reducing the travel distance by $50.5\%$ compared to the FIFO Baseline. By jointly optimizing building-level capacity allocation and intra-building delivery loops, the proposed model consistently outperforms all baselines. These results demonstrate that coordinated hierarchical optimization across both inter-building and intra-building stages is essential for achieving efficient performance in multi-building indoor delivery scenarios.

The Global CVRP attains the same total cost as the Hierarchical CVRP ($2368.0$), indicating that the two-stage decomposition introduces no optimality loss on this small instance. Its solving time, however, is $6.053$\,s compared with $0.961$\,s for the hierarchical model, approximately $6.3$ times longer, as all room-level nodes and all vehicle tours must be handled within a single branch-and-bound search. This discrepancy suggests that the computational advantage of decomposition may become decisive as the instance size increases.

\begin{table}[t]
\centering
\caption{Comparison of delivery strategies in terms of travel cost}
\label{tab:comparison}
\begin{tabular}{lccccc}
\hline
Strategy & Stage~1 Cost & Stage~2 Cost & Total Cost & $\Delta$ (\%) & Solving Time(s)  \\
\hline
FIFO Baseline & 3200.0 & 1580.0 & 4780.0 & -- &0.001\\
Sequential & 1800.0 & 780.0 & 2580.0 & $-46.0\%$ &0.001\\
Hybrid & 1700.0 & 1160.0 & 2860.0 & $-40.2\%$ &0.477\\
Global CVRP & 1700.0 & 668.0 & 2368.0 & $-50.5\%$ &6.053\\
\textbf{Hierarchical CVRP} & \textbf{1700.0} & \textbf{668.0} & \textbf{2368.0} & $\mathbf{-50.5\%}$ &0.961\\
\hline
\end{tabular}
\end{table}

The above results are obtained under a simplified layout with uniform inter-building distances, and therefore provide a conservative estimate of performance gains. In more realistic scenarios with heterogeneous building spacing and irregular internal layouts, the advantage of the Hierarchical CVRP is expected to be more pronounced.

\paragraph{Scalability on a larger instance.}
To examine this conjecture, a larger instance comprising eight buildings, heterogeneous inter-building distances, $81$ room-level demand nodes is evaluated with the same five strategies. Table~\ref{tab:comparison_large} reports the costs and solving times, with $\Delta$ again measured relative to the FIFO Baseline.

\begin{table}[t]
\centering
\caption{Comparison of delivery strategies on the larger instance (eight buildings, 81 room-level demand nodes)}
\label{tab:comparison_large}
\begin{tabular}{lccccc}
\hline
Strategy & Stage~1 Cost & Stage~2 Cost & Total Cost & $\Delta$ (\%) & Solving Time(s) \\
\hline
FIFO Baseline & 17802.0 & 6646.0 & 24448.0 & -- & 0.005 \\
Sequential & 8114.0 & 2735.0 & 10849.0 & $-55.6\%$ & 0.003 \\
Hybrid & 7916.0 & 4141.0 & 12057.0 & $-50.7\%$ & 0.773 \\
Global CVRP & -- & -- & -- & --  & 1000.000 \\
\textbf{Hierarchical CVRP} & \textbf{7916.0} & \textbf{2523.0} & \textbf{10439.0} & $\mathbf{-57.3\%}$ & \textbf{2.256} \\
\hline
\end{tabular}
\end{table}

On this larger instance, the Hierarchical CVRP again yields the lowest total cost ($10439.0$, a $57.3\%$ reduction relative to the FIFO Baseline) in $2.256$\,s. 
Relative to the best heuristic, the Sequential Baseline, this corresponds to a further $3.8\%$ reduction in total cost ($10849.0 \rightarrow 10439.0$) obtained at an absolute computational cost of $2.256$\,s, which remains well within a practical time budget. 
By contrast, the Global CVRP returns no feasible solution within the prescribed $1000$\,s time limit: because the monolithic formulation must simultaneously assign all room-level demands to capacity-constrained tours and eliminate subtours, the solver cannot construct a single feasible incumbent, so that neither a cost value nor the corresponding $\Delta$ can be reported for this strategy. 
These results indicate that the two-stage decomposition provides a strictly better solution than the heuristic baselines within an acceptable solving time, while remaining tractable at a scale where the monolithic model ceases to be solvable; its advantage is thus expected to become more pronounced under heterogeneous building spacing and larger internal layouts.

\subsubsection{Comparison with Static Elevator Cost Modeling}

\begin{figure}
    \centering
    \includegraphics[width=\linewidth]{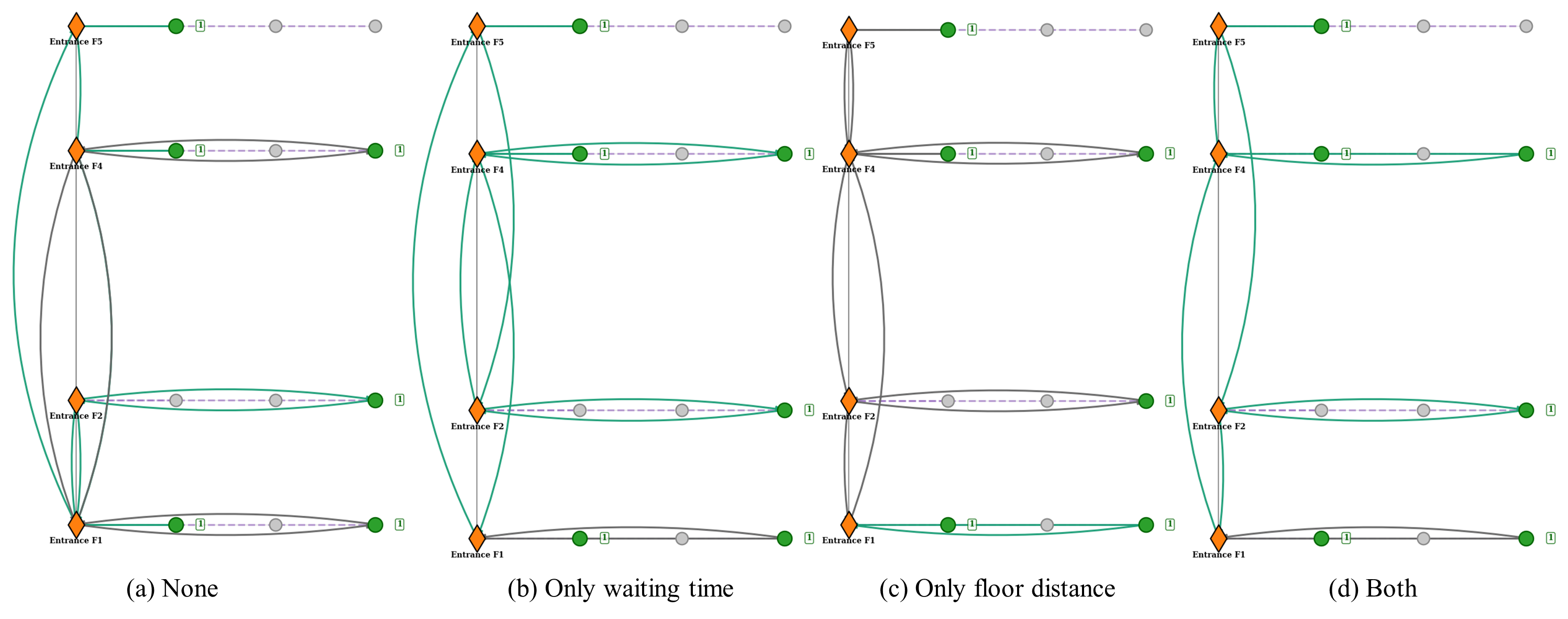}
    \caption{
    Comparison of elevator cost modeling strategies:  
    (a) no elevator modeling,  
    (b) waiting-time–only modeling,  
    (c) floor-distance–only modeling, and  
    (d) the proposed combined model.}
    \label{fig:elevator}
\end{figure}

Figure~\ref{fig:elevator} compares four elevator cost formulations. It is noted that all cases have the same physical distance cost; the difference mainly lies in the vertical transfer.
The baseline without elevator modeling (a) results in excessive and unnecessary vertical transitions, leading to the highest frequency of elevator usage.
Modeling only waiting time (b) suppresses elevator usage but still produces avoidable floor changes due to its lack of spatial awareness.
Modeling only floor distance (c) encourages layer-by-layer routing but may incur excessive elevator time because it ignores waiting delays.
In contrast, the combined formulation (d), which incorporates both waiting time and floor-distance contributions, most accurately captures operational reality and consistently produces the most efficient and rational vertical-transition decisions.

\subsubsection{Comparison with Different Capacity $Q$}

\begin{figure}
    \centering
    \includegraphics[width=\linewidth]{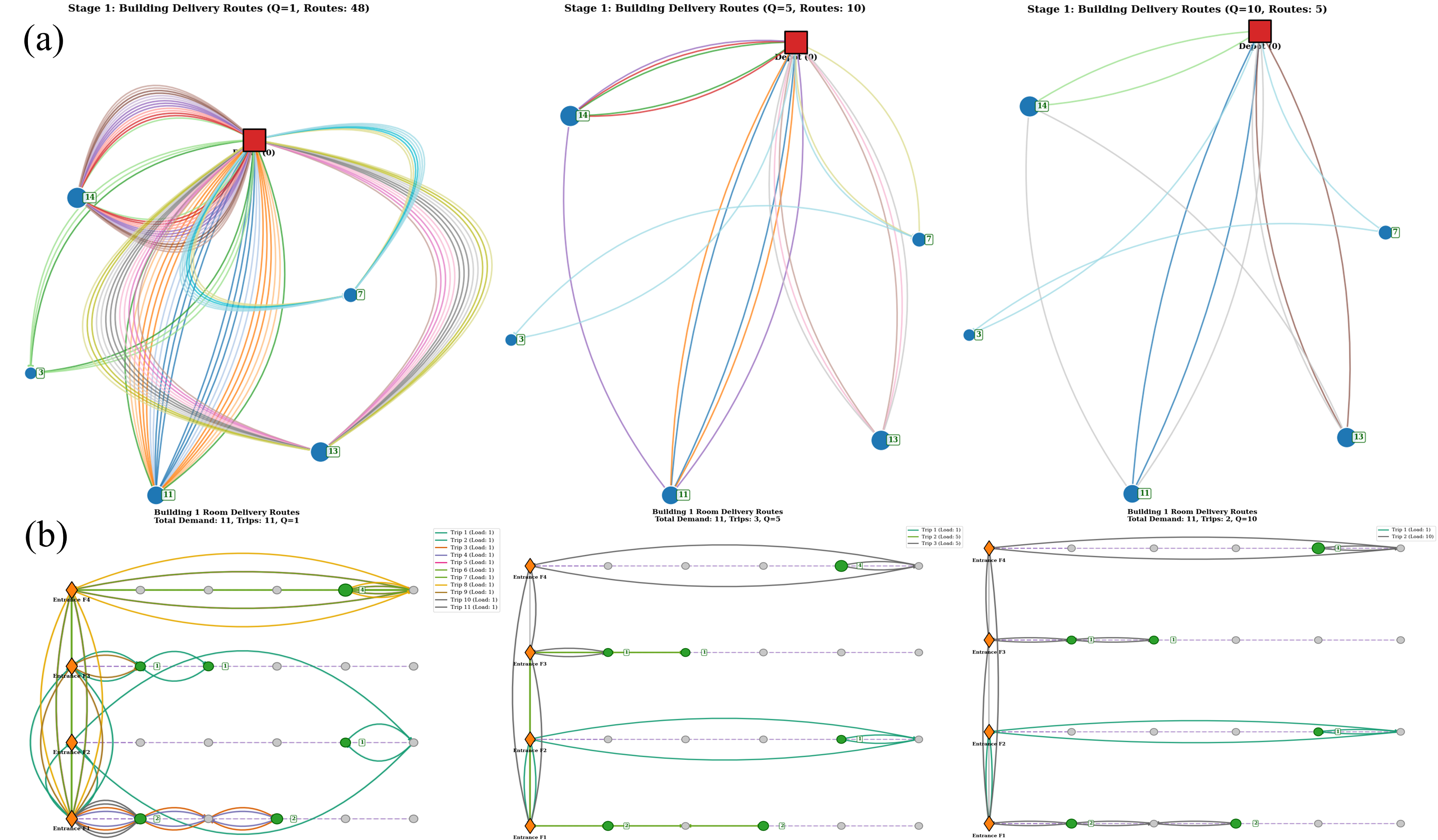}
    \caption{
    Impact of vehicle capacity on scheduling performance.  
    (a) Stage 1: 5-building environment with $Q\in\{1,5,10\}$.  
    (b) Stage 2: single-building environment with $Q\in\{1,5,10\}$.}
    \label{fig:df_Q}
\end{figure}

The influence of vehicle capacity is illustrated in Fig.~\ref{fig:df_Q}.  
Larger capacities reduce the number of required trips and shift the distribution of hierarchical travel costs. When $Q=1$, then every demand package needs a single route.


\subsubsection{Comparison with Scheduling Without Privacy Constraints}

\begin{figure}[htbp]
    \centering
    \includegraphics[width=0.8\linewidth]{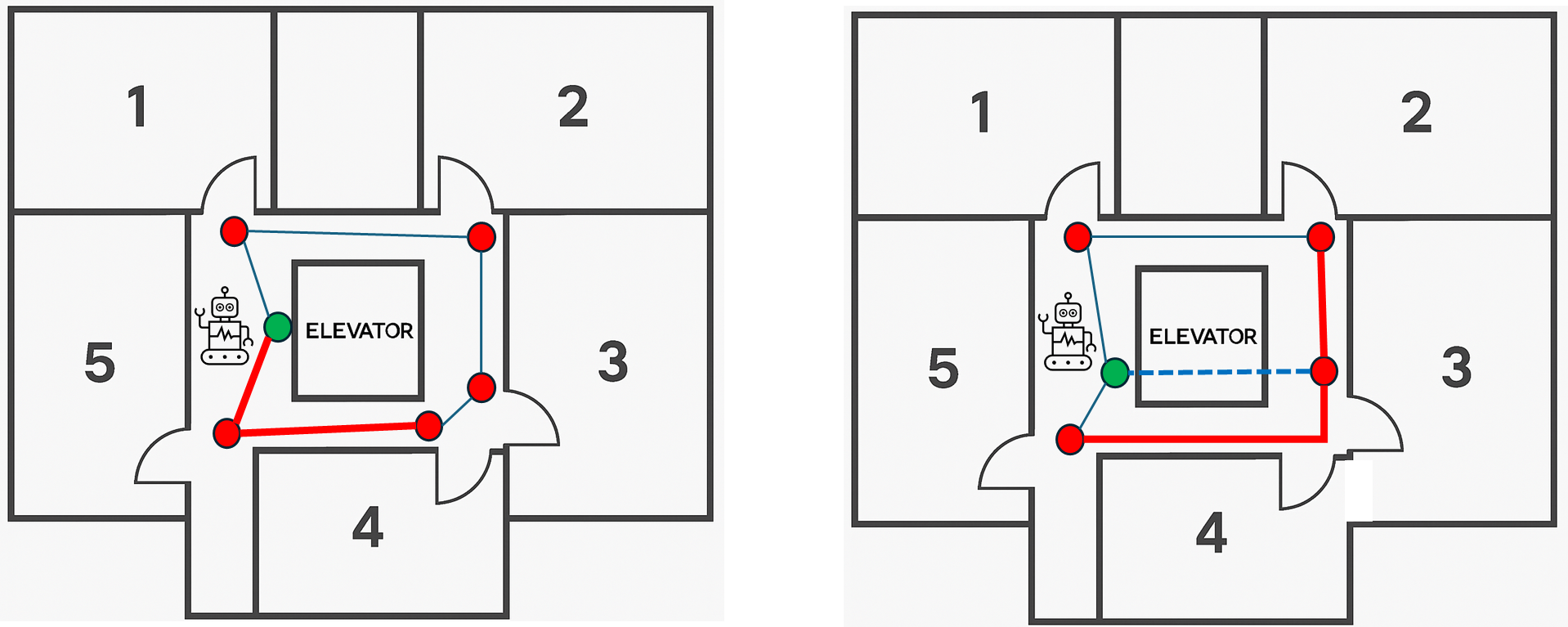}
    \caption{
    Effect of privacy-aware routing on Room-Level navigation.  
    Without privacy penalties, the planner generates paths that enter sensitive residential areas.  
    With privacy-aware cost inflation, such intrusions are avoided.  
    Red line denote privacy-penalized edges.}
    \label{fig:privacy}
\end{figure}

\begin{figure}[htbp]
    \centering
    \includegraphics[width=\linewidth]{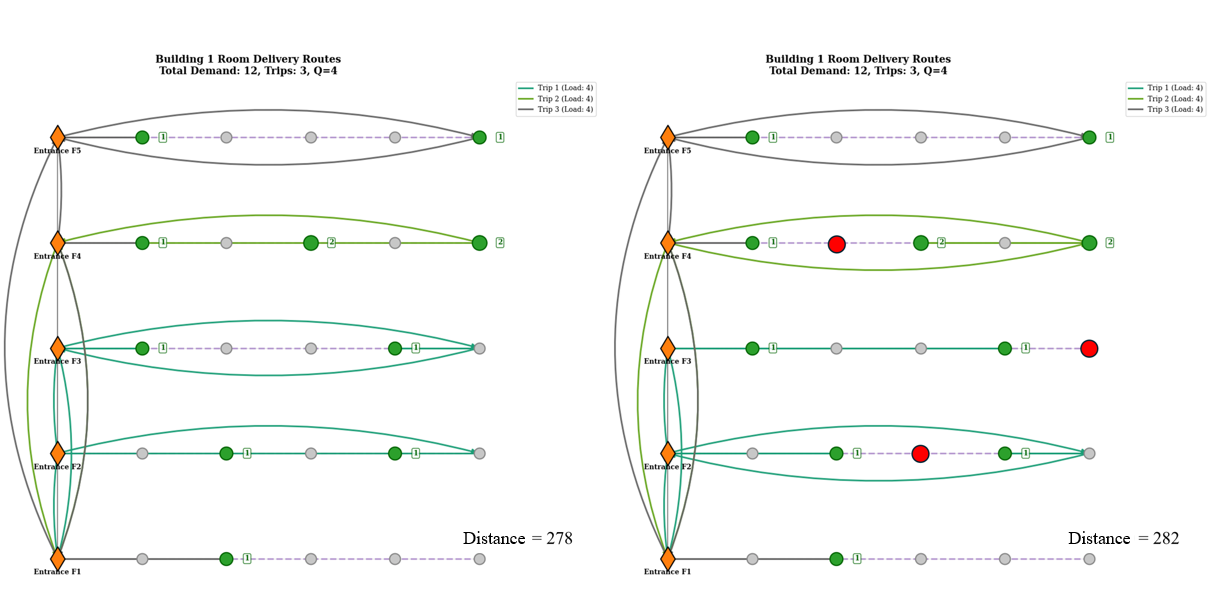}
    \caption{
    Comparison of scheduling outcomes with and without privacy-aware constraints.  
    Left: without privacy penalties;  
    Right: with privacy-aware penalties.  
    Red regions represent privacy-sensitive areas.}
    \label{fig:df_privacy}
\end{figure}

Figures~\ref{fig:privacy} and~\ref{fig:df_privacy} compare the results with and without privacy-aware penalties. Let ${\lambda = 100}$ to represent a large penalty for selected nodes. Privacy constraints clearly prevent the robot from entering restricted corridors, but introduce a slight cost in inter-room distance cost.

Without privacy modeling, planned routes frequently traverse socially sensitive zones.  
Incorporating privacy-aware costs prevents such behaviors, significantly improving deployment feasibility and user acceptance.

\subsection{Computational Efficiency}

\begin{figure}[htbp]
    \centering
    \includegraphics[width=0.8\linewidth]{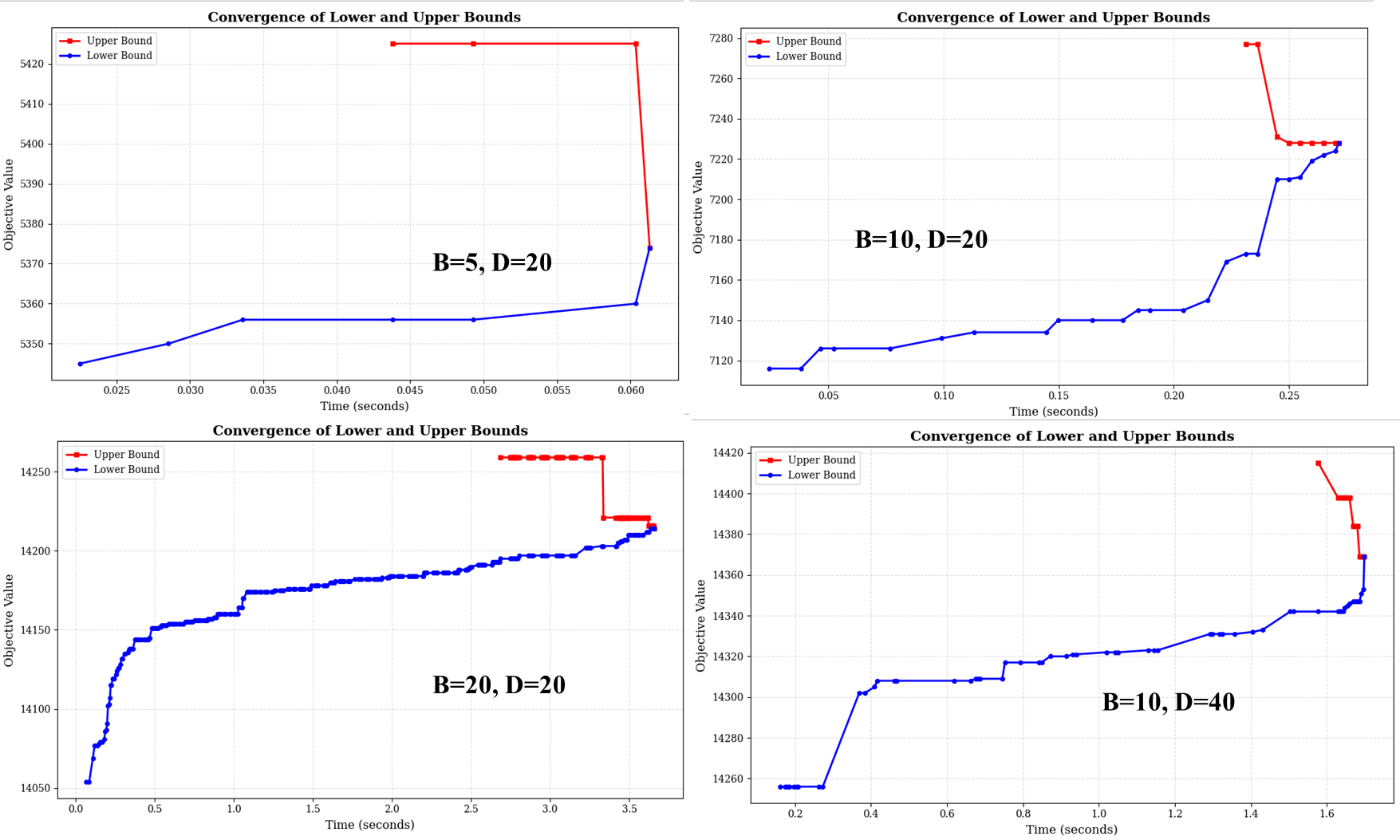}
    \caption{Iteration histories of upper and lower bounds under heterogeneous inter-building distances for the four $(B, D)$ cases.}
    \label{fig:differntdistance}
\end{figure}

\begin{figure}[htbp]
    \centering
    \includegraphics[width=0.8\linewidth]{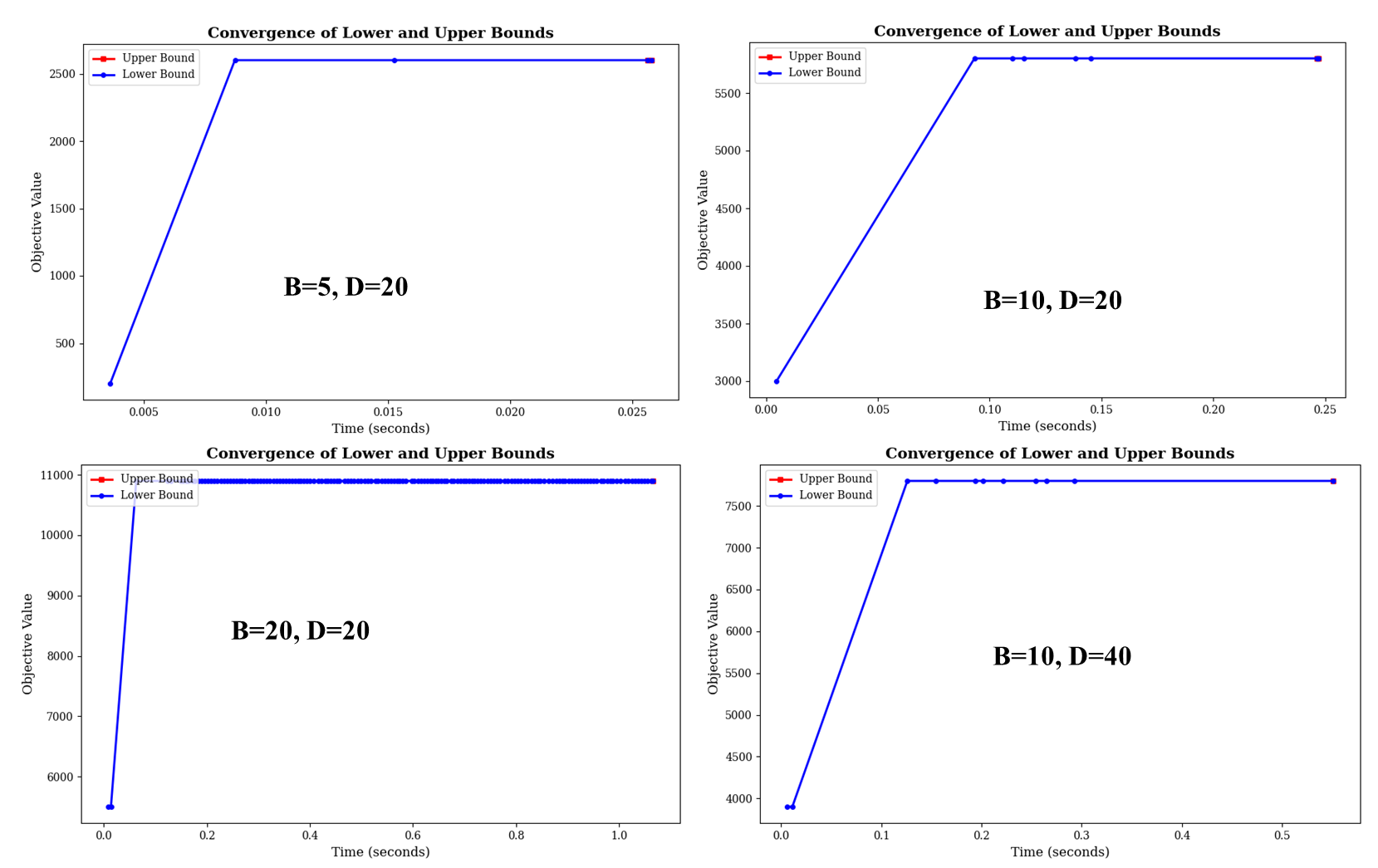}
    \caption{Iteration histories under uniform inter-building distances.}
    \label{fig:fixdistance}
\end{figure}

This section compares the computational efficiency of the proposed CVRP formulation under different problem scales and distance--cost settings. We evaluate four scenarios characterized by the number of buildings $B$ and the maximum per-building demand $D$: 
\textit{Case 1}: $B=5$, $D=20$; 
\textit{Case 2}: $B=10$, $D=20$; 
\textit{Case 3}: $B=20$, $D=20$; 
\textit{Case 4}: $B=10$, $D=40$. 
The optimization trajectories of the upper and lower bounds are presented in Fig.~\ref{fig:differntdistance} (different inter-building distances range from 100 to 200) and Fig.~\ref{fig:fixdistance} (uniform inter-building distances of 100).

All experiments are executed on a compact micro-computer equipped with an \textbf{AMD Ryzen 7 8745H} processor (16 cores). Its small physical footprint and portability make it representative of the compute resources typically available for on-site robotic deployment. Despite the limited size, the solver maintains stable performance across all evaluated scenarios.

As $B$ and $D$ increase, the computational burden grows significantly due to the enlarged combinatorial search space and the increased number of feasibility-enforcing lazy cuts. Under the fixed-distance setting, the lower bound progresses in coarse but steady increments. In contrast, with heterogeneous inter-building distances, uneven edge weights introduce additional complexity into the branch-and-cut process, requiring more iterations for the upper and lower bounds to close and thus leading to slower convergence.

%%%%%%%%%%%%%%%%%%%%%%%%%%%%%%%%%%%%%%%%%%%%%%%%%%%%%%%%%%%%%%%%%%%%%%%%%%%%%%%%

\subsection{Practical Deployment}
\label{subsec:deployment}

The complete autonomous delivery system was deployed and validated through multiple real-world delivery missions spanning several buildings and floors.

\begin{figure}[htbp]
    \centering
    \includegraphics[width=1\linewidth]{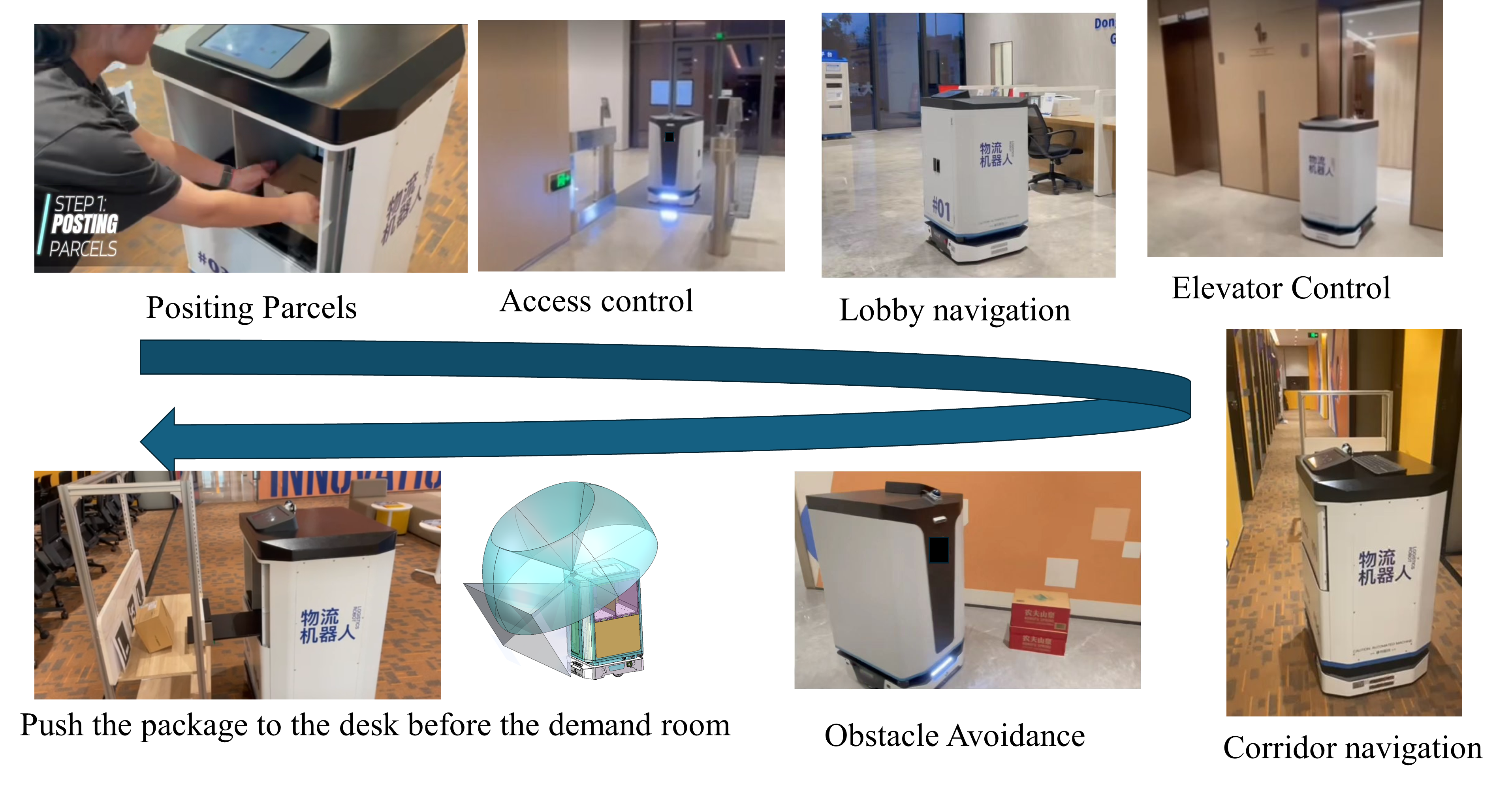}
    \caption{Representative real-world deployment of the autonomous delivery system}
    \label{fig:deploy}
\end{figure}

Fig.~\ref{fig:deploy} illustrates a representative deployment scenario. Each mission begins at a centralized depot, where parcels are loaded into the robot's onboard locker. Guided by a CVRP-optimized delivery plan, the robot autonomously navigates to the target building. Upon arrival, the robot interacts with building infrastructure, including access control systems and elevators, to reach the designated floor. During operation, the robot performs continuous localization and perception, enabling safe navigation and real-time avoidance of both static obstacles and dynamic human traffic. Upon reaching the target room, the robot executes the final delivery by actuating the locker mechanism to release the parcel.

Field deployment results demonstrate the practical robustness and reliability of the proposed system, including reliable elevator interaction with accurate floor detection and consistent re-localization across vertical transitions, robust corridor navigation that enables smooth and collision-free motion in dynamic environments with human presence, and consistent task execution under hierarchical constraints, validating the seamless integration between high-level route planning and low-level motion control while respecting privacy and access constraints.

% \subsubsection{Depots Location Optimization}

% \subsubsection{Comparison with Euclidean-Distance VRP Scheduling}

% Replacing SLAM-derived reachability-aware distances with Euclidean distances resulted in unrealistic routing behavior, including infeasible shortcuts, incorrect elevator choices, and large execution-time detours. These findings confirm that Euclidean distances are poor surrogates for indoor navigation costs.

% Together, these results confirm both the technical soundness and practical deployability of the proposed system in complex multi-building architectural environments.

%%%%%%%%%%%%%%%%%%%%%%%%%%%%%%%%%%%%%%%%%%%%%%%%%%%%%%%%%%%%%%%%%%%%%%%%%%%%%%%%

\section{Conclusion}

This work presents a hierarchical optimization framework for autonomous delivery in large-scale buildings based on a Building--Floor--Room abstraction. By decomposing the original routing problem across multiple spatial levels, the proposed formulation significantly reduces optimization complexity while preserving global feasibility and consistency. A building-aware extension of the capacitated vehicle routing problem (CVRP) is introduced to explicitly model inter-building connectivity, inter-floor transitions, room-level access constraints, and privacy-sensitive regions. The hierarchical model further provides a unified analytical basis for evaluating delivery efficiency under different building layouts, and enables consistent optimization of room-level access strategies.

To demonstrate practical realizability, the proposed algorithms are instantiated in a complete autonomous delivery system that tightly integrates hierarchical planning with mapping, localization, and navigation. A non-interactive, privacy-preserving delivery robot and reachability-aware distance metrics derived from real building geometry ensure consistency between theoretical planning results and physical execution. Extensive multi-building and multi-floor experiments confirm that the algorithmic advantages translate into reliable real-world performance, effectively bridging abstract optimization models and deployable autonomous delivery systems.

Future work will explore theoretical performance bounds for the hierarchical routing formulation, extension to multi-robot cooperative optimization, and uncertainty-aware planning under dynamic demand and access constraints, further strengthening both the theoretical foundations and practical scalability of hierarchical autonomous delivery.

%% ==========================================================================
%% Back-matter sections required by the ESWA template (B/template.tex).
%% The original manuscript did not contain these statements; fill in the
%% placeholders before submission.
%% ==========================================================================

\section*{Declaration of Competing Interest}
% Standard declaration taken from B/template.tex; please confirm it applies.
The authors declare that they have no known competing financial interests or personal relationships that could have appeared to influence the work reported in this paper.

\bibliography{reference}

\end{document}